\documentclass[runningheads]{llncs}
\usepackage{graphicx}
\usepackage{amsmath,amssymb} 
\usepackage{color}

\usepackage{multirow}
\usepackage{csquotes}
\usepackage{xcolor}
\usepackage{makecell}
\usepackage{blindtext}
\usepackage{pifont}
\usepackage{tabularx} 
\usepackage{capt-of}
\usepackage[export]{adjustbox}
\usepackage{hyperref}

\DeclareMathOperator*{\argmin}{arg\,min}

\newcommand{\cmark}{\ding{51}}%
\newcommand{\xmark}{\ding{55}}%

\newcolumntype{L}[1]{>{\raggedright\let\newline\\\arraybackslash\hspace{0pt}}m{#1}}

\definecolor{greengt}{RGB}{106,168,79}

\begin{document}
\title{How to Read Paintings: Semantic Art Understanding with Multi-Modal Retrieval} 

\titlerunning{How to Read Paintings}
%
\author{Noa Garcia \and George Vogiatzis}
%
\authorrunning{N. Garcia and G. Vogiatzis}
%

\institute{Aston University, United Kingdom \\
\email{\{garciadn,g.vogiatzis\}@aston.ac.uk}}
\maketitle              
\begin{abstract}
Automatic art analysis has been mostly focused on classifying artworks into different artistic styles. However, understanding an artistic representation involves more complex processes, such as identifying the elements in the scene or recognizing author influences. We present SemArt, a multi-modal dataset for semantic art understanding. SemArt is a collection of fine-art painting images in which each image is associated to a number of attributes and a textual artistic comment, such as those that appear in art catalogues or museum collections. To evaluate semantic art understanding, we envisage the Text2Art challenge, a multi-modal retrieval task where relevant paintings are retrieved according to an artistic text, and vice versa. We also propose several models for encoding visual and textual artistic representations into a common semantic space. Our best approach is able to find the correct image within the top 10 ranked images in the 45.5\% of the test samples. Moreover, our models show remarkable levels of art understanding when compared against human evaluation.
\end{abstract}

\section{Introduction}
The ultimate aim of computer vision has always been to enable computers to understand images the way humans do. With the latest advances in deep learning technologies, the availability of large volumes of training data and the use of powerful graphic processing units, computer vision systems are now able to locate and classify objects in natural images with high accuracy, surpassing human performance in some specific tasks. However, we are still a long way from human-like analysis and extraction of high-level semantics from images. This work aims to push high-level image recognition by enabling machines to interpret art. 

\begin{figure}[t]
\centering
\setlength{\tabcolsep}{2pt}
\begin{tabular}[t]{c c}
\includegraphics[width=0.45\textwidth]{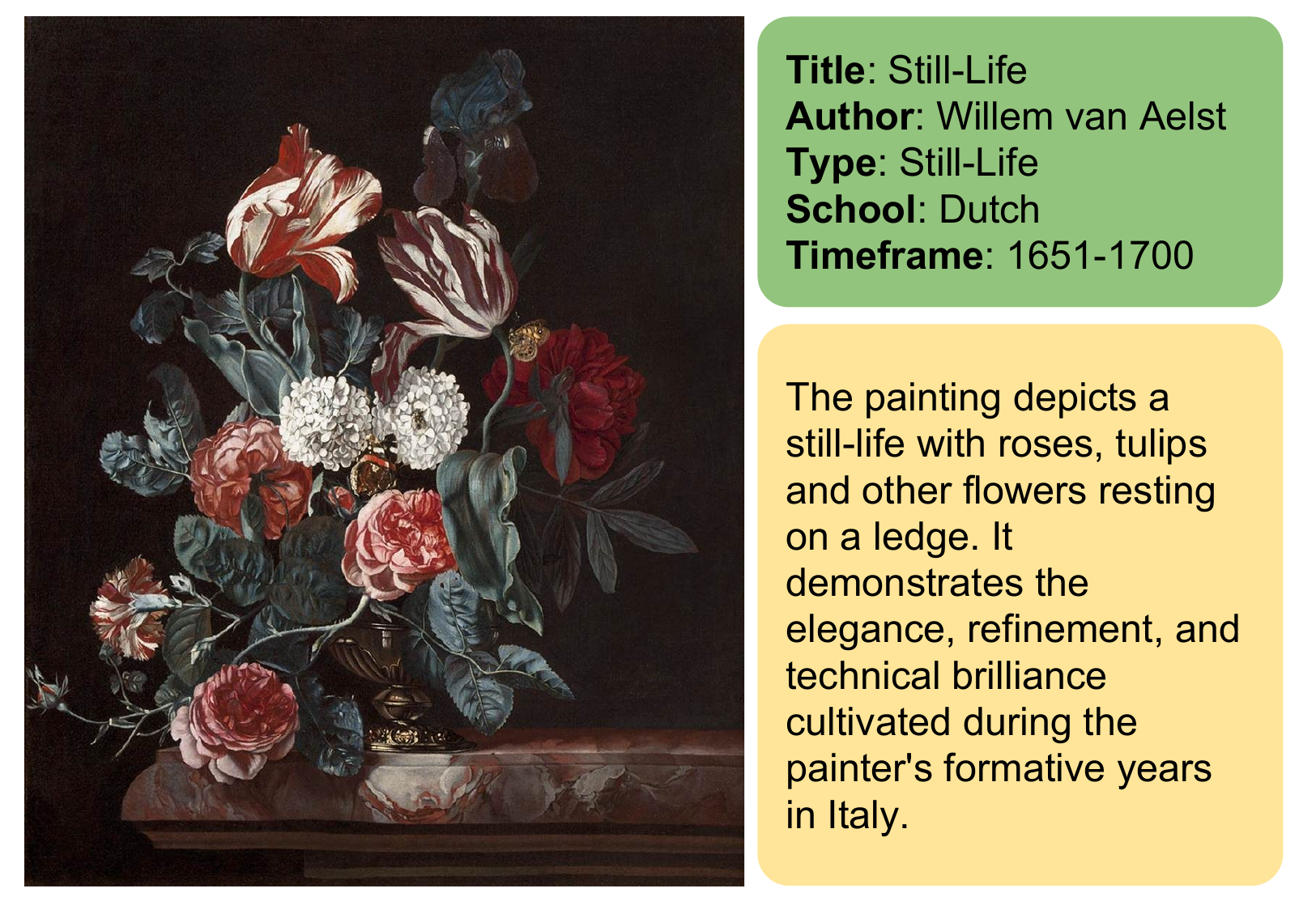} & 
\includegraphics[width=0.45\textwidth]{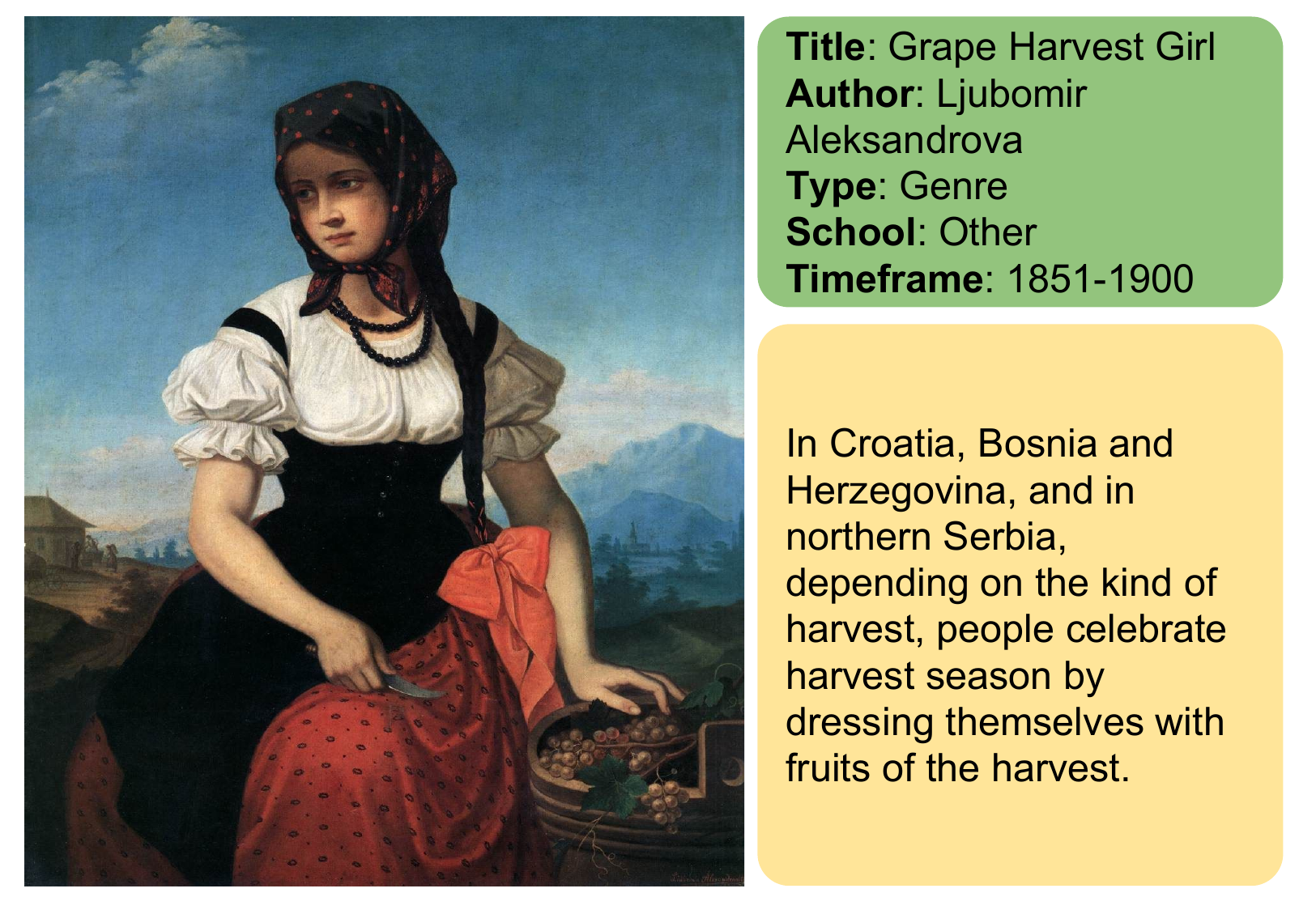}
\end{tabular}
\caption{\textbf{SemArt dataset samples}. Each sample is a triplet of image, attributes and artistic comment.}
\label{fig:example}
\end{figure}

To study automatic interpretation of art, we introduce SemArt\footnote{\href{http://noagarciad.com/SemArt/}{http://noagarciad.com/SemArt/}}, a dataset for semantic art understanding. We build SemArt by gathering a collection of fine-art images, each with its respective attributes (author, type, school, etc.) as well as a short artistic comment or description, such as those that commonly appear in art catalogues or museum collections. Artistic comments  involve not only descriptions of the visual elements that appear in the scene but also references to its technique, author or context. Some examples of the dataset are shown in Figure \ref{fig:example}.

We address semantic art understanding by proposing a number of models that map paintings and artistic comments into a common semantic space, thus enabling comparison in terms of semantic similarity. To evaluate and benchmark the proposed models, we design the Text2Art challenge as a multi-modal retrieval task. The aim of the challenge is to evaluate whether the models capture enough of the insights and clues provided by the artistic description to be able to match it to the correct painting. 

A key difference with previously proposed methods for semantic understanding of natural images (e.g. MS-COCO dataset \cite{Lin2014}) is that our system relies on background information on art history and artistic styles. As already noted in previous work \cite{crowley2014state,crowley2014search,crowley2015face}, paintings are substantially different from natural images in several aspects. Firstly, paintings, unlike natural images, are figurative representations of people, objects, places or situations which may or may not correspond to the real world. Secondly, the study of fine-art paintings usually requires previous knowledge about history of art, different artistic styles as well as contextual information about the subjects represented. Thirdly, paintings commonly exhibit one or more layers of abstraction and symbolism which creates ambiguity in interpretation. 

In this work, we harness existing prior knowledge about art and deep neural networks to model understanding of fine-art paintings. Specifically, our contributions are: 
\begin{enumerate}
\item to introduce the first dataset for semantic art understanding in which each sample is a triplet of images, attributes and artistic comments,
\item to propose models to map fine-art paintings and their high-level artistic descriptions onto a joint semantic space,
\item to design an evaluation protocol based on multi-modal retrieval for semantic art understanding, so that future research can be benchmarked under a common, public framework.
\end{enumerate}

\section{Related Work}
With the digitalization of large collections of fine-art paintings and the emergence of publicly available online art catalogs such as WikiArt\footnote{http://www.wikiart.org} or the Web Gallery of Art\footnote{https://www.wga.hu/}, computer vision researchers become interested in analyzing fine-art paintings automatically. Early work \cite{johnson2008image,shamir2010impressionism,carneiro2012artistic,khan2014painting} proposes methods based on handcrafted visual features to identify an author and/or a specific style in a piece of art. Datasets used in these kinds of approaches, such as PRINTART \cite{carneiro2012artistic} and  Painting-91 \cite{khan2014painting}, are rather small, with 988 and 4,266 painting images, respectively. Mensink and Van Gemert introduce in \cite{mensink2014rijksmuseum} the large-scale Rijksmuseum dataset for multi-class prediction, consisting on 112,039 images from artistic objects, although only 3,593 are from fine-art paintings. With the success of convolutional neural networks (CNN) in large-scale image classification \cite{krizhevsky2012imagenet}, deep features from CNNs replace handcrafted image representations in many computer vision applications, including painting image classification \cite{Bar2014ClassificationOA,karayev2014recognizing,Saleh2015LargescaleCO,Tan2016CeciNP,ma2017part,mao2017deepart}, and larger datasets are made publicly available \cite{karayev2014recognizing,mao2017deepart}. In these methods, paintings are fed into a CNN to predict its artistic style or author by studying its visual aesthetics. 

Besides painting classification, other work is focused on exploring image retrieval in artistic paintings. For example, in \cite{carneiro2012artistic}, monochromatic painting images are retrieved by using artistic-related keywords, whereas in \cite{seguin2016visual} a pre-trained CNN is fine-tuned to find paintings with similar artistic motifs.  Crowley and Zisserman \cite{crowley2015face} explore domain transfer to retrieve image of portraits from real faces, in the same way as \cite{crowley2014state} and \cite{crowley2016art} explore domain transfer to perform object recognition in paintings. 

\begin{table}[t]
\setlength{\tabcolsep}{5pt}
\caption{\textbf{Datasets for art analysis}. \textit{Meta} and \textit{Text} columns state if image metadata and textual information are provided, respectively.}
\centering
\begin{tabular}{ l c c c l}
\Xhline{2\arrayrulewidth}
\textbf{Dataset} & \textbf{\#Paintings} & \textbf{Meta} & \textbf{Text} & \textbf{Task}  \\ \hline 
PRINTART \cite{carneiro2012artistic} & 988 & \cmark & \xmark & Classification and Retrieval \\
Painting-91 \cite{khan2014painting} & 4,266 & \cmark & \xmark & Classification \\
Rijksmuseum \cite{mensink2014rijksmuseum} & 3,593 & \cmark & \xmark &  Classification \\
Wikipaintings \cite{karayev2014recognizing} & 85,000 &  \cmark & \xmark & Classification\\
Paintings \cite{crowley2014state} & 8,629 & \xmark & \xmark & Object Recognition \\
Face Paintings \cite{crowley2015face} & 14,000 & \xmark & \xmark & Face Retrieval \\ 
VisualLink \cite{seguin2016visual} & 38,500 & \cmark & \xmark & Instance Retrieval \\
Art500k \cite{mao2017deepart} & 554,198 & \cmark & \xmark & Classification \\
SemArt & 21,383 & \cmark & \cmark & Semantic Retrieval \\
\Xhline{2\arrayrulewidth}
\end{tabular}
\label{tab:artdatasets}
\end{table}

A summary of the existing datasets for fine-art understanding is shown in Table \ref{tab:artdatasets}. In essence, previous work studies art from an aesthetics point of view to classify paintings according to author and style \cite{carneiro2012artistic,khan2014painting,mensink2014rijksmuseum,karayev2014recognizing,mao2017deepart}, to find relevant paintings according to a query input \cite{carneiro2012artistic,crowley2015face,seguin2016visual} or to identify objects in artistic representations \cite{crowley2014state}. However, understanding art involves also identifying the symbolism of the elements, the artistic influences or the historical context of the work. To study such complex processes, we propose to interpret fine-art paintings in a semantic way by introducing SemArt, a multi-modal dataset for semantic art understanding. To the best of our knowledge, SemArt is the first corpus that provides not only fine-art images and their attributes, but also artistic comments for the semantic understanding of fine-art paintings.

\section{SemArt Dataset}
\label{sec:dataset}
\subsection{Data Collection}
To create the SemArt dataset, we collect artistic data from the Web Gallery of Art (WGA), a website with more than 44,809 images of European fine-art reproductions between the 8th and the 19th century. WGA provides links to all their images in a downloadable comma separated values file (CSV). In the CSV file, each image is associated with some attributes or metadata: author, author's birth and death, title, date, technique, current location, form, type, school and time-line. Following the links provided in the CSV file, we only collect images from artworks whose field \textit{form} is set as painting, as opposite to images of other forms of art such as sculpture or architecture.

We create a script to collect artistic comments for each painting image, as they are not provided in the aforementioned CSV file. We omit images that are not associated to any comment and we remove irrelevant metadata fields, such as author's birth and death and current location. The final size of the cleaned collection is downsampled to 21,384 triplets, where each triplet is formed by an image, a text and a number of attributes. 

\subsection{Data Analysis}
\begin{figure}[t]
\centering
\setlength{\tabcolsep}{-12pt}
\begin{tabular}{c c c }
\includegraphics[width = 0.41\textwidth]{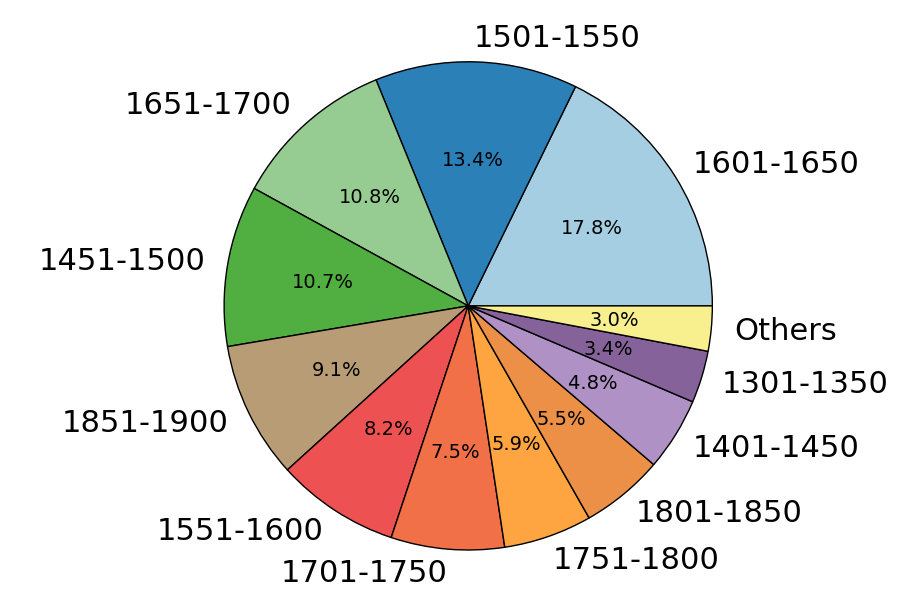} &
\includegraphics[width = 0.41\textwidth]{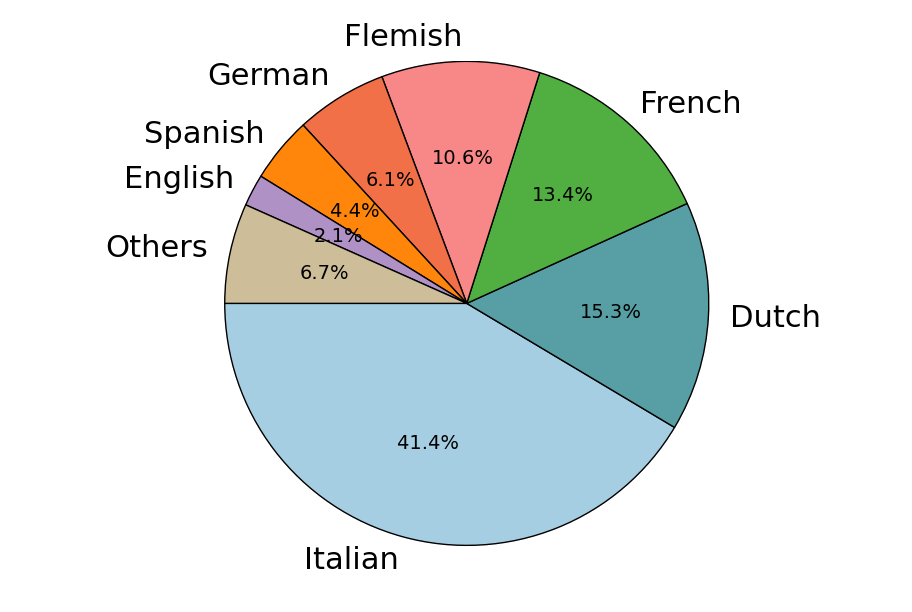} &
\includegraphics[width = 0.41\textwidth]{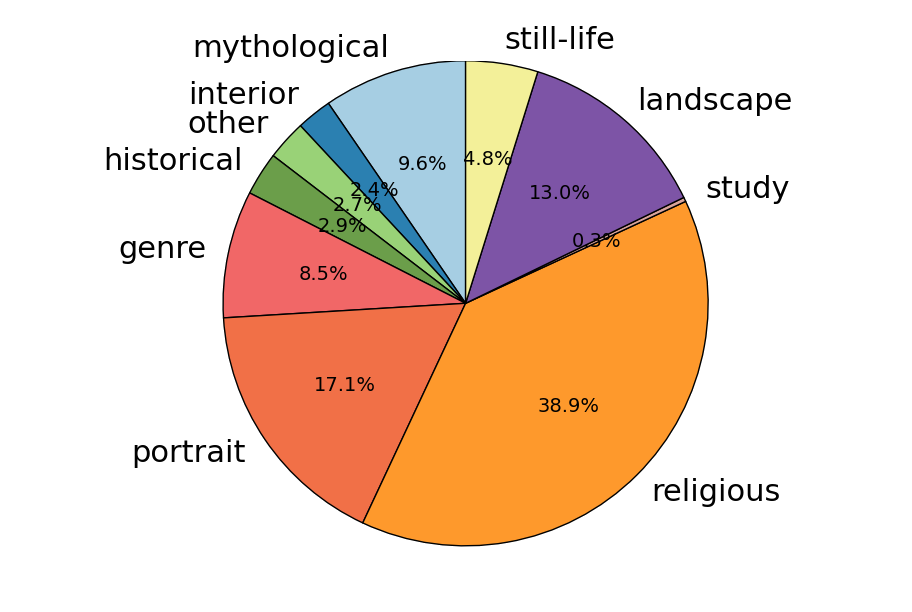} \\[5pt]
Timeframe & School & Type \\
\end{tabular}
\caption{\textbf{Metadata distribution}. Distribution of samples within the SemArt dataset in Timeframe, School and Type attributes.}
\label{fig:piecharts}
\end{figure}

For each sample, the metadata is provided as a set of seven fields, which describe the basic attributes of its associated painting: \textit{Author}, \textit{Title}, \textit{Date}, \textit{Technique}, \textit{Type}, \textit{School} and \textit{Timeframe}. In total, there are 3,281 different authors, the most frequent one being Vincent van Gogh with 327 paintings. There are 14,902 different titles in the dataset, with 38.8\% of the paintings presenting a non-unique title. Among all the titles, Still-Life and Self-Portrait are the most common ones. \textit{Technique} and \textit{Date} fields are not available for all samples, but provided for completeness. \textit{Type} field classifies paintings according to ten different genres, such as religious, landscape or portrait. There are 26 artistic schools in the collection, Italian being the most common, with 8,860 paintings and Finnish the least frequent with just 5 samples. Also, there are 22 different timeframes, which are periods of 50 years evenly distributed between 801 and 1900. The distribution of values over the fields \textit{Type}, \textit{School} and \textit{Timeframe} is shown in Figure \ref{fig:piecharts}. With respect to artistic comments, the vocabulary set follows the Zipf's law \cite{Manning2001FoundationsOS}. Most of the comments are relatively short, with almost 70\% of the them containing 100 words or less. Images are provided in different aspect ratios and sizes. The dataset is randomly split into training, validation and test sets with 19,244, 1,069 and 1,069 triplets, respectively. 

\section{Text2Art Challenge}
\label{sec:retrieval}
In what follows, we use bold style to refer to vectors and matrices (e.g $\vec{x}$ and $\vec{W}$). Given a collection of artistic samples $K$, the $k$-th sample in $K$ is given by the triplet $(img_k, com_k, att_k)$, being $img_k$ the artistic image, $com_k$ the artistic comment and $att_k$ the artistic attributes. Images, comments and attributes are input into specific encoding functions, $f_{img}$, $f_{com}$, $f_{att}$, to map raw data from the corpus into vector representations, $\vec{i}_k$, $\vec{c}_k$, $\vec{a}_k$, as: %
\begin{equation}
\vec{i}_k = f_{img}(img_k;\phi_{img})
\end{equation} %
\begin{equation}
\vec{c}_k = f_{com}(com_k;\phi_{com})
\end{equation} %
\begin{equation}
\vec{a}_k = f_{att}(att_k;\phi_{att})
\end{equation} %
where $\phi_{img}$, $\phi_{com}$ and $\phi_{att}$ are the parameters of each encoding function. 

As comment encodings, $\vec{c}_k$, and attribute encodings, $\vec{a}_k$, are both from textual data, a joint textual vector, $\vec{t}_k$ can be obtained as: %
\begin{equation}
\vec{t}_k = \vec{c}_k \oplus \vec{a}_k
\end{equation} %
where $\oplus$ is vector concatenation.

The transformation functions, $g_{vis}$ and $g_{text}$, can be defined as the functions that project the visual and the textual encodings into a common multi-modal space. The projected vectors $\vec{p}^{vis}_k$ and $\vec{p}^{text}_k$ are then obtained as: %
\begin{equation}
\vec{p}^{vis}_k = g_{vis}(\vec{i}_k;\theta_{vis})
\end{equation} %
\begin{equation}
\vec{p}^{text}_k = g_{text}(\vec{t}_k;\theta_{text})
\end{equation} %
being $\theta_{vis}$ and $\theta_{text}$ the parameters of each transformation function. 

For a given similarity function $d$, the similarity between any text (i.e. pair of comments and attributes) and any image in $K$ is measured as the distance between their projections: \begin{equation}
d(\vec{p}^{text}_k, \vec{p}^{vis}_j) = d(g_{text}(\vec{t}_k;\theta_{text}), g_{vis}(\vec{i}_j;\theta_{vis}))
\end{equation}

In semantic art understanding, the aim is to learn $f_{img}$, $f_{com}$, $f_{att}$, $g_{vis}$ and $g_{text}$ such that 
images, comments and attributes from the same sample are mapped closer in terms of $d$ than images, texts and attributes from different samples: \begin{equation}
d(\vec{p}^{text}_k, \vec{p}^{vis}_k) < d(\vec{p}^{text}_k, \vec{p}^{vis}_j) \text{ for all } k, j \leq |K|
\end{equation} and \begin{equation}
d(\vec{p}^{text}_k, \vec{p}^{vis}_k) < d(\vec{p}^{text}_j, \vec{p}^{vis}_k) \text{ for all } k, j \leq |K|
\end{equation}

To evaluate semantic art understanding, we propose the Text2Art challenge as a multi-modal retrieval problem. Within Text2Art, we define two tasks: text-to-image retrieval and image-to-text retrieval. In text-to-image retrieval, the aim is to find the most relevant painting in the collection, $img^* \in K$, given a query comment and its attributes: %
\begin{equation}
img^* = \argmin_{img_j \in K} d(\vec{p}^{text}_k, \vec{p}^{vis}_j)
\end{equation} %

Similarly, in the image-to-text retrieval task, when a painting image is given, the aim is to find the comment and the attributes, $com^* \in K$ and $att^* \in K$ , that are more relevant to the visual query: %
\begin{equation}
com^*, att^* = \argmin_{com_j, att_j \in K} d(\vec{p}^{text}_j, \vec{p}^{vis}_k)
\end{equation}

\section{Models for Semantic Art Understanding}
\label{sec:method}
We propose several models to learn meaningful textual and visual encodings and transformations for semantic art understanding. First, images, comments and attributes are encoded into visual and textual vectors. Then, a multi-modal transformation model is used to map these visual and textual vectors into a common multi-modal space where a similarity function  is applied.

\subsection{Visual Encoding}
We represent each painting image as a visual vector, $\vec{i}_k$, using convolutional neural networks (CNNs). We use different CNN architectures, such as VGG16 \cite{Simonyan15}, different versions of ResNet \cite{he2016deep} and RMAC \cite{tolias2015particular}. %

\begin{description}
\item[VGG16] \cite{Simonyan15} contains 13 3x3 convolutional layers and three fully-connected layers stacked on top of each other. We use the output of one of the fully connected layers as the visual encoding.
\item[ResNet] \cite{he2016deep} uses shortcut connections to connect the input of a layer to the output of a deeper layer. There exist many versions depending on the number of layers, such as ResNet50 and ResNet152 with 50 and 152 layers, respectively. We use the output of the last layer as the visual encoding.
\item[RMAC] is a visual descriptor introduced by Tolias et al. in \cite{tolias2015particular} for image retrieval. The activation map from the last convolutional layer from a CNN model is max-pooled over several regions to obtain a set of regional features. The regional features are post-processed, sum-up together and normalized to obtain the final visual representation.
\end{description}

\subsection{Textual Encoding}
With respect to the textual information, comments are encoded into a comment vector, $\vec{c}_k$, and attributes are encoded into an attribute vector, $\vec{a}_k$. To get the joint textual encoding, $\vec{t}_k$, both vectors are concatenated.

\subsubsection{Comment Encoding}
\label{sec:textencoding}
To encode comments into a comment vector, $\vec{c}_k$, we first build a comment vocabulary, $V_C$. $V_C$ contains all the alphabetic words that appear at least ten times in the training set. The comment vector is obtained using three different techniques: a comment bag-of-words (BOW\scriptsize{c}\normalsize), a comment multi-layer perceptron (MLP\scriptsize{c}\normalsize) and a comment recurrent model (LSTM\scriptsize{c}\normalsize). %
\begin{description}
\item[BOW\scriptsize{c}] each comment is encoded as a \textit{term frequency - inverse document frequency} (tf-idf) vector by weighting each word in the comment by its relevance within the corpus.
\item[MLP\scriptsize{c}] comments are encoded as a tf-idf vectors and fed into a fully connected layer with tanh activation\footnote{$\text{tanh}(z) = \frac{e^z - e^{-z}} {e^z + e^{-z}}$} and $\ell_2$-normalization. The output of the normalization layer is used as the comment encoding.
\item[LSTM\scriptsize{c}] each sentence in a comment is encoded into a sentence vector using a 2,400 dimensional pre-trained skip-thought model \cite{kiros2015skip}. Sentence vectors are input into a long short-term memory network (LSTM) \cite{hochreiter1997long}. The last state of the LSTM is $\ell_2$-normalized and used as the comment encoding.
\end{description}

\subsubsection{Attribute Encoding}
We use the attribute field \textit{Title} in the metadata to provide an extra textual information to our model. We propose three different techniques to encode titles into attribute encodings, $\vec{a}_k$: an attribute bag-of-words (BOW\scriptsize{a}\normalsize) an attribute multi-layer perceptron (MLP\scriptsize{a}\normalsize) and an attribute recurrent model (LSTM\scriptsize{a}\normalsize).

\begin{description}
\item[BOW\scriptsize{a}] as in comments, titles are encoded as a tf-idf-weighted vector using a title vocabulary, $V_T$. $V_T$ is built with all the alphabetic words in the titles of the training set.
\item[MLP\scriptsize{a}] also as in comments, tf-idf encoded titles are fed into a fully connected layer with tanh activation and a $\ell_2$-normalization. The output of the normalization layer is used as the attribute vector.
\item[LSTM\scriptsize{a}] in this case, each word in a title is fed into an embedding layer followed by a LSTM network. The output of the last state of the LSTM is $\ell_2$-normalized and used as the attribute encoding. 
\end{description}

\subsection{Multi-Modal Transformation}
\begin{figure}[t]
\centering
\begin{tabular}{c c c }
\textbf{CCA Model} & \textbf{CML Model} & \textbf{AMD Model} \\[5pt]
\includegraphics[height = 0.35\textwidth]{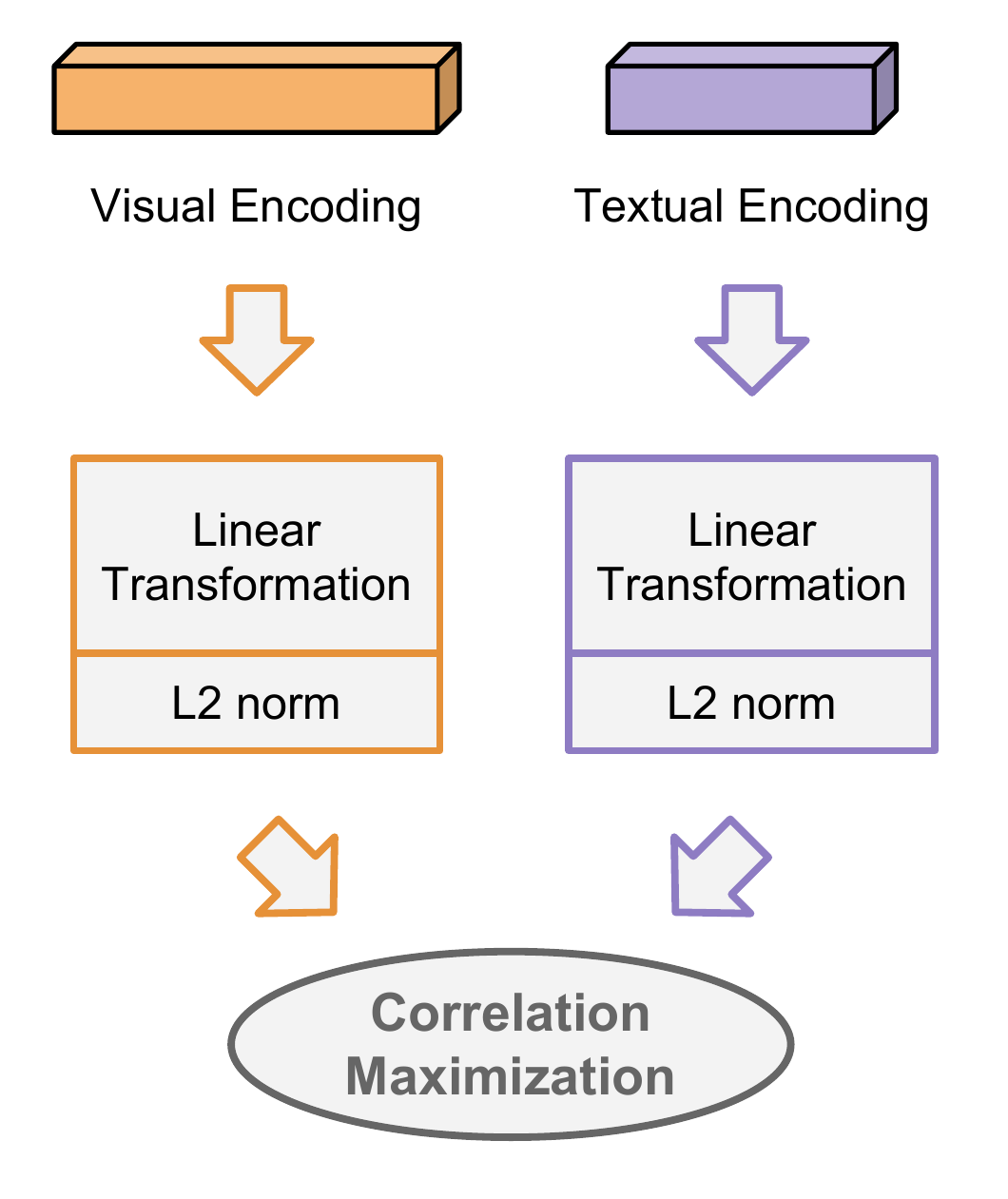} &
\includegraphics[height = 0.35\textwidth]{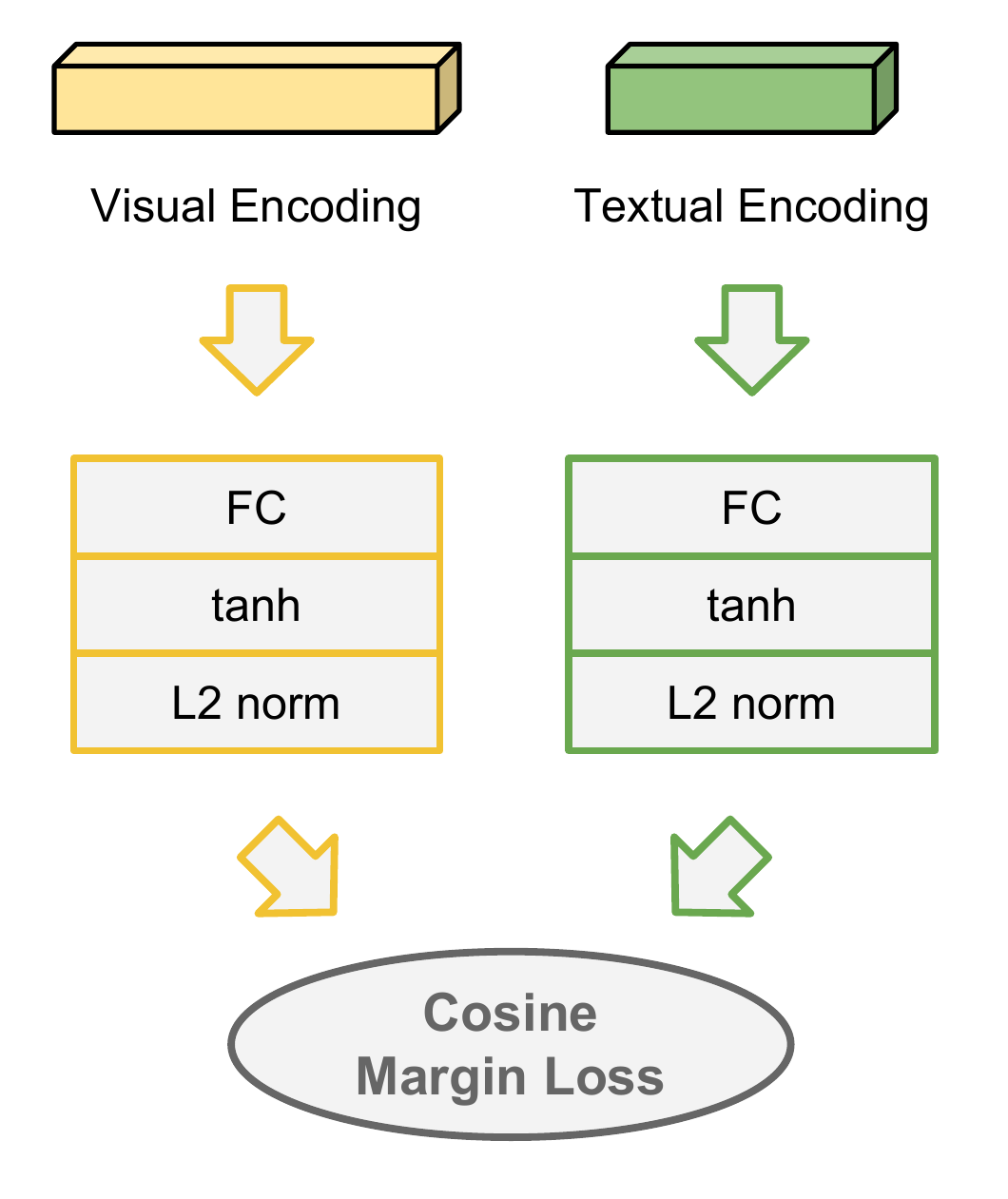} &
\includegraphics[height = 0.35\textwidth]{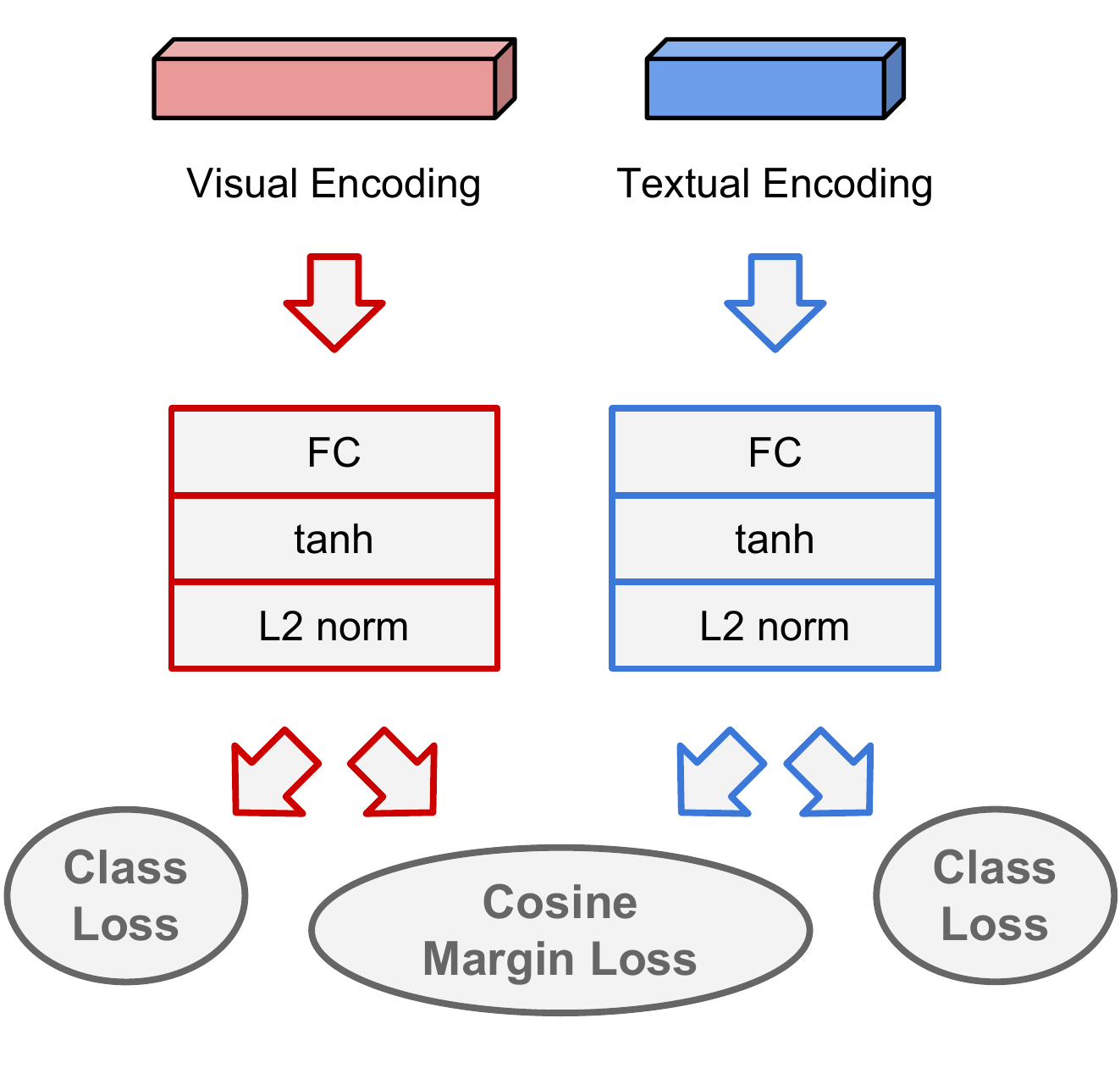} \\
\end{tabular}
\caption{\textbf{Multi-modal transformation models}. Models for mapping textual and visual representations into a common multi-modal space.}
\label{fig:threemodels}
\end{figure}

The visual and textual encodings, $\vec{i}_k$ and $\vec{t}_k$ respectively, encode visual and textual data into two different spaces. We use a multi-modal transformation model to map the visual and textual representations into a common multi-modal space. In this common space, textual and visual information can be compared in terms of the similarity function $d$. We propose three different models, which are illustrated in Figure \ref{fig:threemodels}.

\begin{description}
\item[CCA] Canonical Correlation Analysis (CCA) \cite{gong2014multi} is a linear approach for projecting data from two different sources into a common space by maximizing the normalized correlation between the projected data. The CCA projection matrices are learnt by using training pairs of samples from the corpus. At test time, the textual and visual encodings from a test sample are projected using these CCA matrices.

\item[CML] Cosine Margin Loss (CML) is a deep learning architecture trained end-to-end to learn the visual and textual encodings and their projections all at once. Each image encoding is fed into a fully connected layer followed by a tanh activation function and a $\ell_2$-normalization layer to project the visual feature, $\vec{i}_j$, into a $D$-dimensional space, obtaining the projected visual vector $\vec{p}^{vis}_j$. Similarly, each textual vector $\vec{t}_k$, is input into another network with identical layer structure (fully connected layer with tanh activation and $\ell_2$-normalization) to map the textual feature into the same $D$-dimensional space, obtaining the projected textual vector $\vec{p}^{text}_k$. We train the CML model with both positive ($k = j$) and negative ($k \neq j$) pairs of textual and visual data and cosine similarity with margin as the loss function:  \begin{equation}
\begin{split}L_{CML}(\vec{p}^{vis}_k, \vec{p}^{text}_j) =
\begin{cases}
1 - \cos(\vec{p}^{vis}_k, \vec{p}^{text}_j), & \text{if } k = j \\
\max(0, \cos(\vec{p}^{vis}_k, \vec{p}^{text}_j) - m), & \text{if } k \neq j
\end{cases}\end{split}
\end{equation} where $\text{cos}$ is the cosine similarity between two normalized vectors and $m$ is the margin hyperparameter.

\item[AMD] Augmented Metadata (AMD) is a model in which the network is informed with attribute data for an extra alignment between the visual and the textual encodings. The AMD model consists on a deep learning architecture that projects both visual and textual vectors into the common multi-modal space whereas, at the same time, ensures that the projected encodings are meaningful in the art domain. As in the CML model, image and textual encodings are projected into $D$-dimensional vectors using fully connected layers, and the loss between the multi-modal transformations is computed using a cosine margin loss. Attribute metadata is used to train a pair of classifiers on top of the projected data (Figure \ref{fig:threemodels}, AMD Model), each classifier consisting of a fully connected layer without activation. Metadata classifiers are trained using a standard cross entropy classification loss function: \begin{equation}
L_{META}(\vec{x}, class) = -\log\left(\frac{\exp(\vec{x}[class])}{\sum_j \exp(\vec{x}[j])}\right)
\end{equation} which contribute to the total loss of the model in addition to the cosine margin loss. The total loss of the model is then computed as:
\begin{equation}
\begin{split}
L_{AMD}(\vec{p}^{text}_k, \vec{p}^{vis}_j,l_{p^{text}_k}, l_{p^{vis}_j}) = (1 - 2\alpha) L_{CML}(\vec{p}^{text}_k, \vec{p}^{vis}_j)\\ 
+ \alpha L_{META}(\vec{p}^{text}_k, l_{p^{text}_k}) \\
+ \alpha L_{META}(\vec{p}^{vis}_j, l_{p^{vis}_j})
\end{split}
\end{equation} where $l_{p^{text}_k}$ and $l_{p^{vis}_j}$ are the class labels of the $k$-th text and the $j$-th image, respectively, and $\alpha$ is the weight of the classifier loss.
\end{description}

\section{Experiments}
\label{sec:experiments}

\subsubsection{Experimental Details.} In the image encoding part, each network is initialized with its standard pre-trained weights for image classification. Images are scaled down to 256 pixels per side and randomly cropped into $224 \times 224$ patches. Visual data is augmented by randomly flipping images horizontally. In the textual encoding part, the dimensionality of LSTM hidden state for comments is 1,024, whereas in the LSTM for titles is 300. The title vocabulary size is 9,092. Skip thoughts dimensionality is set to 2,400. In the multi-modal transformation part, the CCA matrices are learnt using scikit-learn \cite{scikit-learn}. For the deep learning architectures, we use Adam optimizer and the learning rate is set to $0.0001$, $m$ to $0.1$ and $\alpha$ to 0.01. Training is conducted in mini batches of $32$ samples. Cosine similarity is used as the similarity function $d$ in all of our models.

\subsubsection{Text2Art Challenge Evaluation.} Painting images are ranked according to their similarity to a given text, and vice versa. The ranking is computed on the whole set of test samples and results are reported as median rank (MR) and recall rate at K (R@K), with K being 1, 5 and 10. MR is the value separating the higher half of the relevant ranking position amount all samples, so the lower the better. Recall at rate K is the rate of samples for which its relevant image is in the top K positions of the ranking, so the higher the better.

\subsection{Visual Domain Adaptation}
We first evaluate the transferability of visual features from the natural image domain to the artistic domain. In this experiment, texts are encoded with the BOW\scriptsize{c} \normalsize approach with $V_C =$ 3,000. As multi-modal transformation model, a 128-dimensional CCA is used. We extract visual encodings from networks pre-trained for classification of natural images without further fine-tunning or refinement.  For the VGG16 model, we extract features from the first, second and third fully connected layer (VGG16\scriptsize{FC1} \normalsize, VGG16\scriptsize{FC2} \normalsize and VGG16\scriptsize{FC3}\normalsize). For the ResNet models, we consider the visual features from the output of the networks (ResNet50 and ResNet152). Finally, RMAC representation is computed using a VGG16, a ResNet50 and a ResNet152 (RMAC\scriptsize{VGG16} \normalsize, RMAC\scriptsize{Res50} \normalsize and RMAC\scriptsize{Res152}\normalsize). Results are detailed in Table \ref{tab:imagefeatures}. As semantic art understanding is a high-level task, it is expected that representations acquired from deeper layers perform better, as in the VGG16 models, where the deepest layer of the network obtains the best performance. RMAC features respond well when transferring from natural images to art, although ResNet models obtain the best performance. Considering these results, we use ResNets as visual encoders in the following experiments.

\begin{table}[t]
\setlength{\tabcolsep}{7pt}
\caption{\textbf{Visual Domain Adaptation.} Transferability of visual features from the natural image classification domain to the Text2Art challenge.}
\centering
\resizebox{\textwidth}{!}{%
\begin{tabular}{ l c c c c c c c c c c c}
\Xhline{2\arrayrulewidth}
\multicolumn{2}{c}{\textbf{Encoding}}  & & \multicolumn{4}{c}{\textbf{Text-to-Image}} & & \multicolumn{4}{c}{\textbf{Image-to-Text}} \\ \cline{1-2} \cline{4-7} \cline{9-12}
\multicolumn{1}{c}{\textbf{Img}} & \textbf{Dim} & & \textbf{R@1} & \textbf{R@5} & \textbf{R@10} & \textbf{MR} & & \textbf{R@1} & \textbf{R@5} & \textbf{R@10} & \textbf{MR} \\ \hline
VGG16 \scriptsize{FC1} & 4,096 & & 0.069 & 0.129 & 0.174 & 115 & & 0.061 & 0.129 & 0.180 & 121 \\
VGG16 \scriptsize{FC2} & 4,096 & & 0.051 & 0.097 & 0.109 & 278 & & 0.051 & 0.085 & 0.103 & 275 \\
VGG16 \scriptsize{FC3} & 1,000 & & 0.101 & 0.211 & 0.285 & 44 & & 0.094 & 0.217 & 0.283 & 51 \\
ResNet50 & 1,000 & & 0.114 & 0.231 & 0.304 & 42 & & 0.114 & 0.242 & 0.318 & 44 \\
ResNet152 & 1,000 & & 0.108 & \textbf{0.254} & \textbf{0.343} & \textbf{36} & & \textbf{0.118} & \textbf{0.250} & \textbf{0.321} & \textbf{36} \\
RMAC \scriptsize{VGG16}& 512 & & 0.092 & 0.206 & 0.286 & 41 & & 0.084 & 0.202 & 0.293 & 44 \\ 
RMAC \scriptsize{Res50} & 2,048 & & 0.084 & 0.202 & 0.293 & 48 & & 0.097 & 0.215 & 0.288 & 49 \\
RMAC \scriptsize{Res152} & 2,048 & & \textbf{0.115} & 0.233 & 0.306 & 44 & & 0.103 & 0.238 & 0.305 & 44   \\ 
\Xhline{2\arrayrulewidth}
\end{tabular}}
\label{tab:imagefeatures}
\end{table}

\begin{table}[t]
\setlength{\tabcolsep}{7pt}
\caption{\textbf{Text Encoding in Art.} Comparison between different text encodings in the Text2Art challenge.}
\centering
\resizebox{\textwidth}{!}{%
\begin{tabular}{ c c c c c c c c c c c c}
\Xhline{2\arrayrulewidth}
\multicolumn{2}{c}{\textbf{Encoding}} & & \multicolumn{4}{c}{\textbf{Text-to-Image}} & & \multicolumn{4}{c}{\textbf{Image-to-Text}} \\ \cline{1-2} \cline{4-7} \cline{9-12}
\textbf{Com} & \textbf{Att} & & \textbf{R@1} & \textbf{R@5} & \textbf{R@10} & \textbf{MR} & & \textbf{R@1} & \textbf{R@5} & \textbf{R@10} & \textbf{MR} \\ \hline

LSTMc & LSTMa & & 0.053 & 0.162 & 0.256 & 33 & & 0.053 & 0.180 & 0.268 & 33 \\
MLPc & LSTMa & & 0.089 & 0.260 & 0.376 & 21 & & 0.093 & 0.249 & 0.363 & 21\\
MLPc & MLPa & & 0.137 & 0.306 & 0.432 & 16 & & \textbf{0.140} & 0.317 & 0.436 & 15 \\
BOWc & BOWa & & \textbf{0.144} & \textbf{0.332} & \textbf{0.454} & \textbf{14} & & 0.138 & \textbf{0.327} & \textbf{0.457} & \textbf{14} \\

\Xhline{2\arrayrulewidth}
\end{tabular}}
\label{tab:textfeatures}
\end{table}

\subsection{Text Encoding in Art}
We then compare the performance between the different text encoding models in the Text2Art challenge. In this experiment, images are encoded with a ResNet50 network and the CML model is used to learn the mapping of the visual and the textual encodings into a common 128-dimensional space. The different encoding methods are compared in Table \ref{tab:textfeatures}. The best performance is obtained when using the simplest bag-of-words approach both for comments and titles (BOW{\scriptsize c} and BOW{\scriptsize a}), although the multi-layer perceptron model (MLP{\scriptsize c} and MLP{\scriptsize a}) obtain similar results. Models based on recurrent networks (LSTM{\scriptsize c} and LSTM{\scriptsize a}) are not able to capture the insights of semantic art understanding. These results are consistent with previous work \cite{wang2018learning}, which shows that text recurrent models perform worse than non-recurrent methods for multi-modal tasks that do not require text generation.

\subsection{Multi-Modal Models for Art Understanding}
\begin{figure}
\setlength{\tabcolsep}{3pt}
\centering
\scriptsize
\resizebox{\textwidth}{!}{%
\begin{tabular}[t]{c c c c c}
\multicolumn{5}{L{12cm}}{\scriptsize{\textbf{Title}: Still-Life of Apples, Pears and Figs in a Wicker Basket on a Stone Ledge \linebreak
\textbf{Comment}: The large dark vine leaves and fruit are back-lit and are sharply silhouetted against the luminous background, to quite dramatic effect. Ponce's use of this effect strongly indicates the indirect influence of Caravaggio's Basket of Fruit in the Pinacoteca Ambrosiana, Milan, almost 50 years after it was created.}} \\
& \\ [-5pt]
\includegraphics[width = 0.18\textwidth, height=60pt, cfbox=greengt 2pt 0pt]{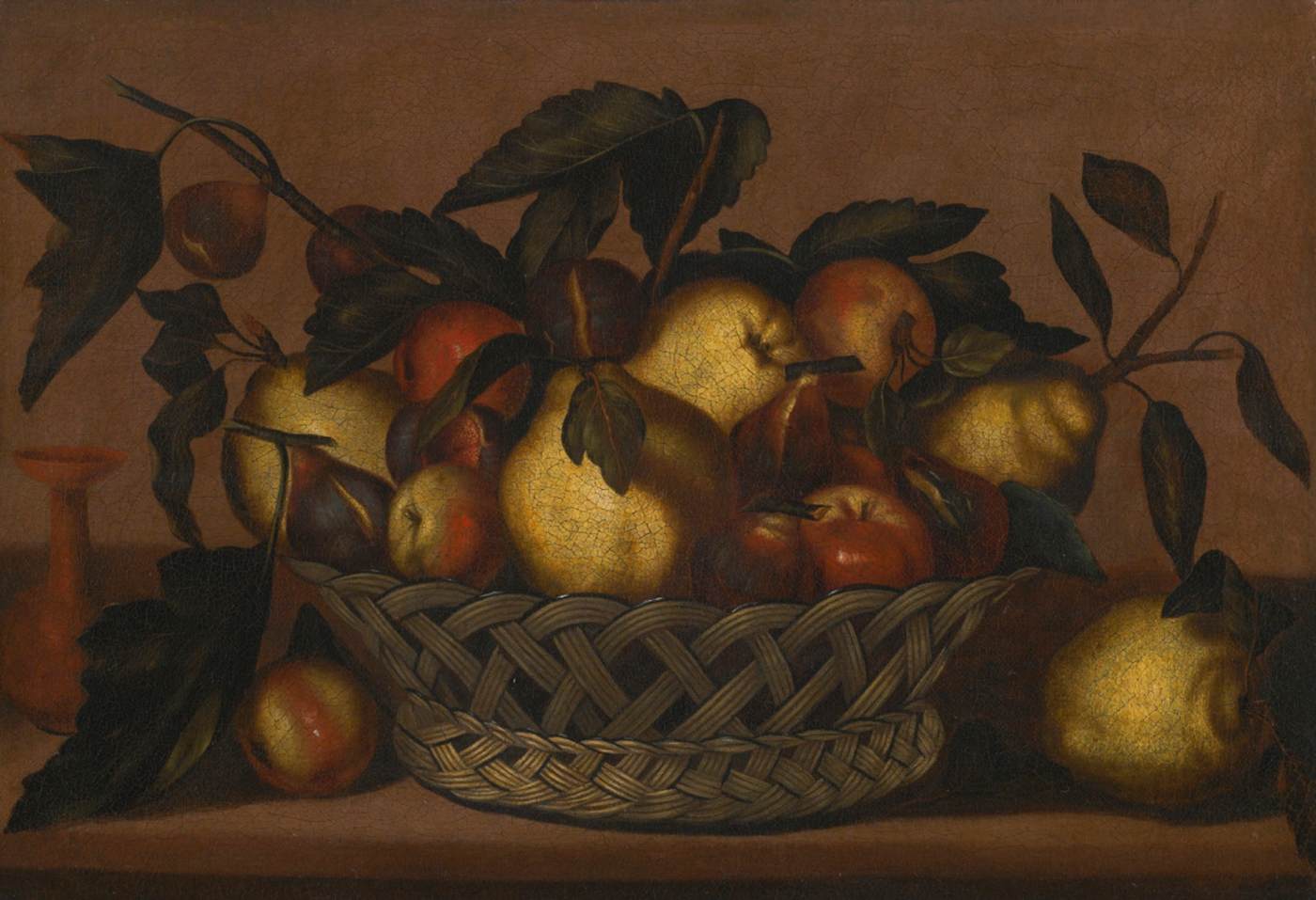} &
\includegraphics[width = 0.18\textwidth, height=60pt]{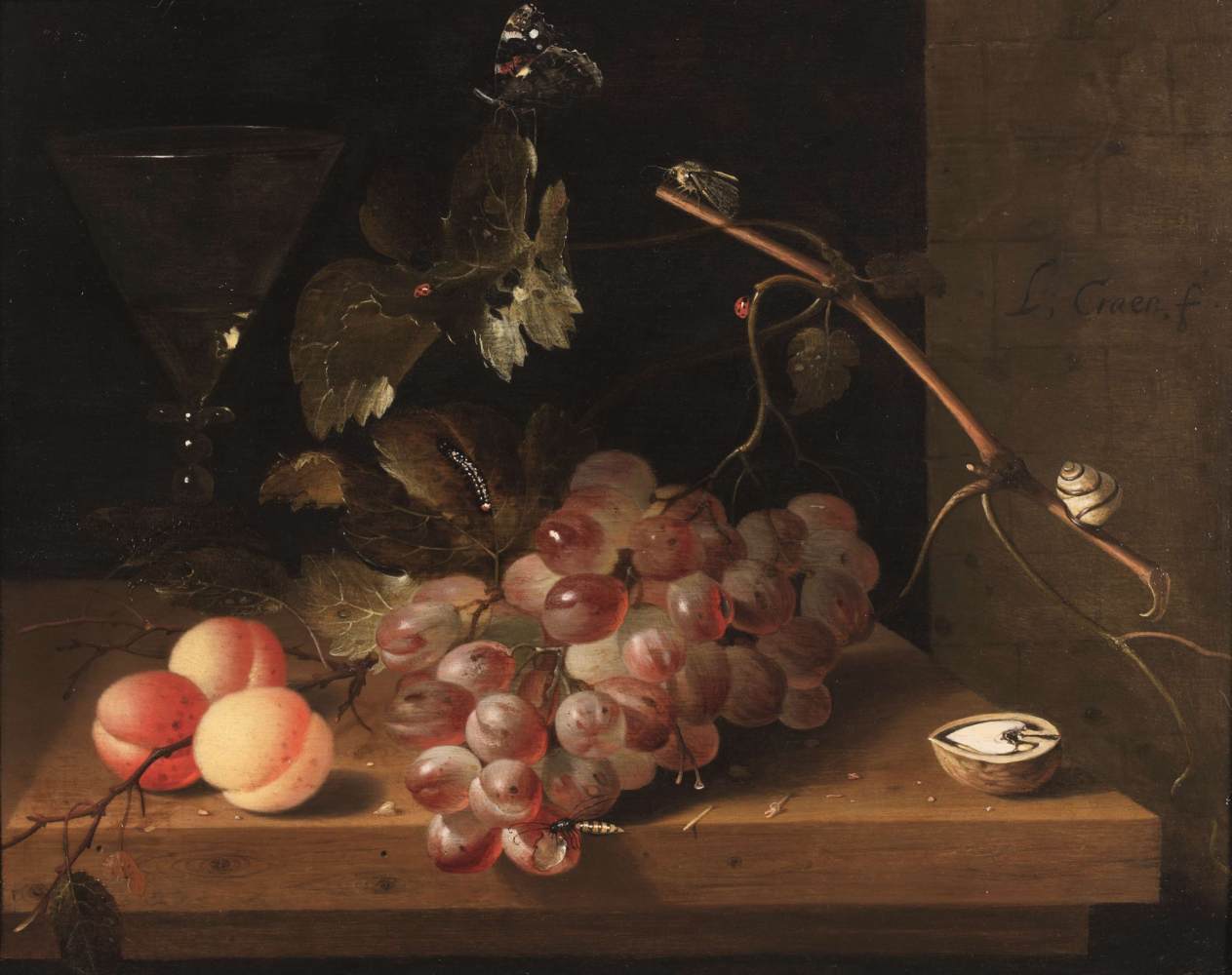} &
\includegraphics[width = 0.18\textwidth, height=60pt]{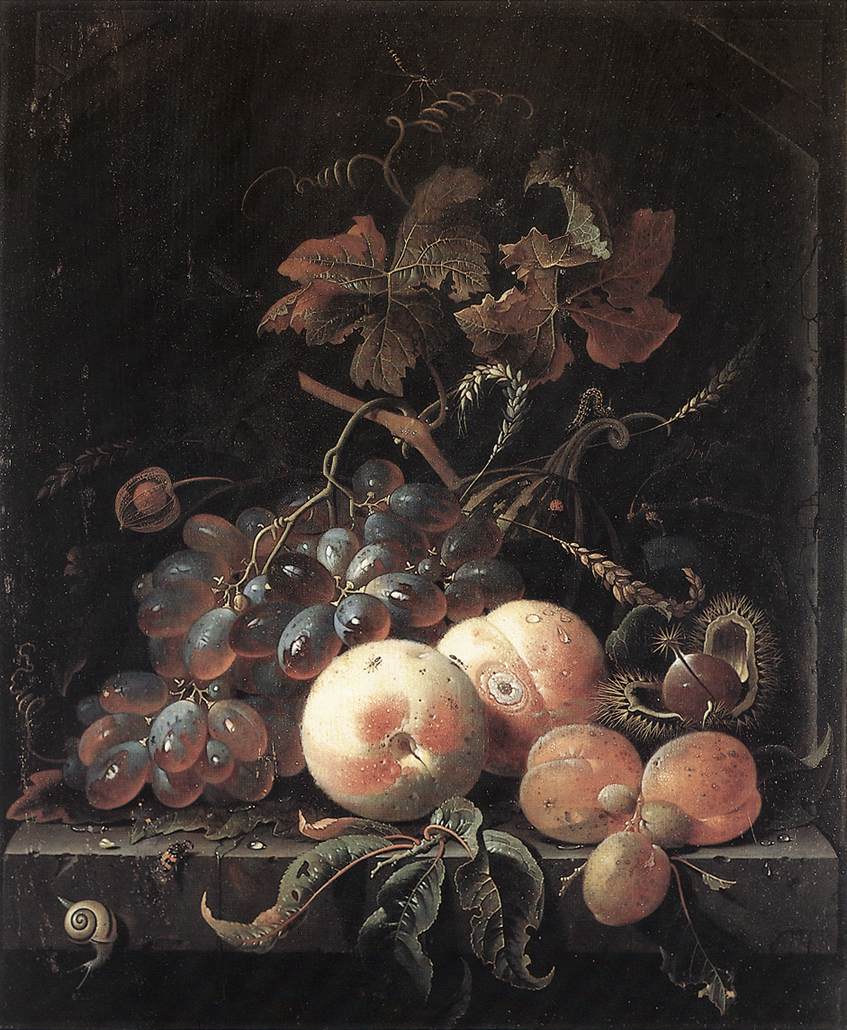} &
\includegraphics[width = 0.18\textwidth, height=60pt]{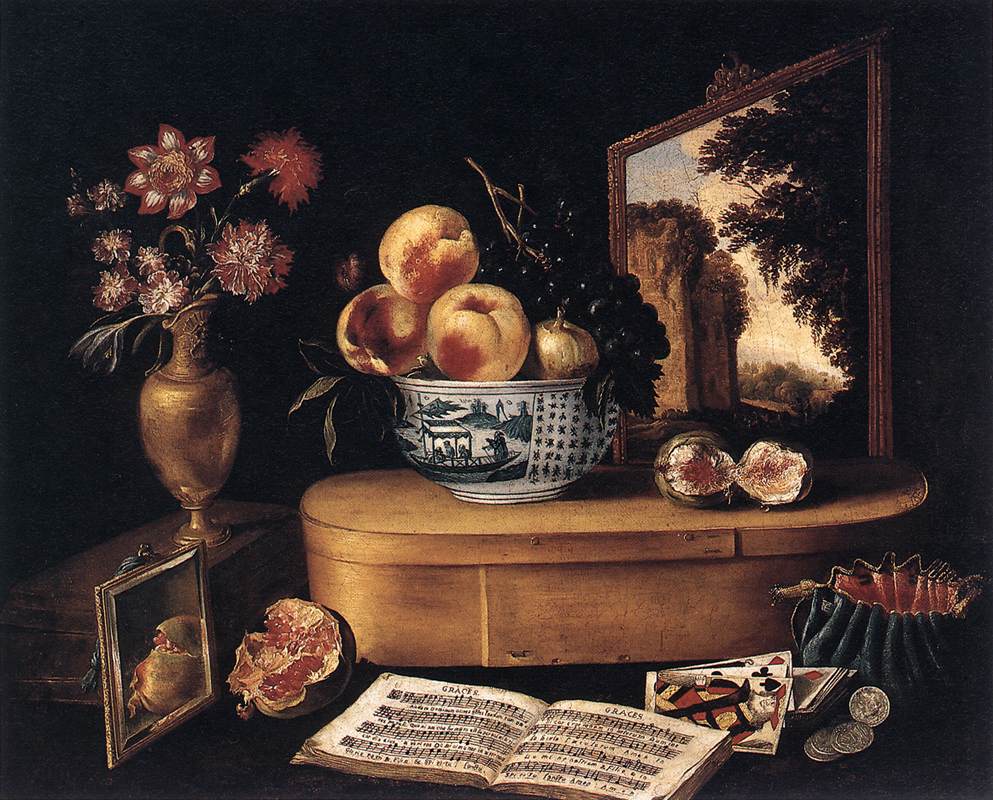} &
\includegraphics[width = 0.18\textwidth, height=60pt]{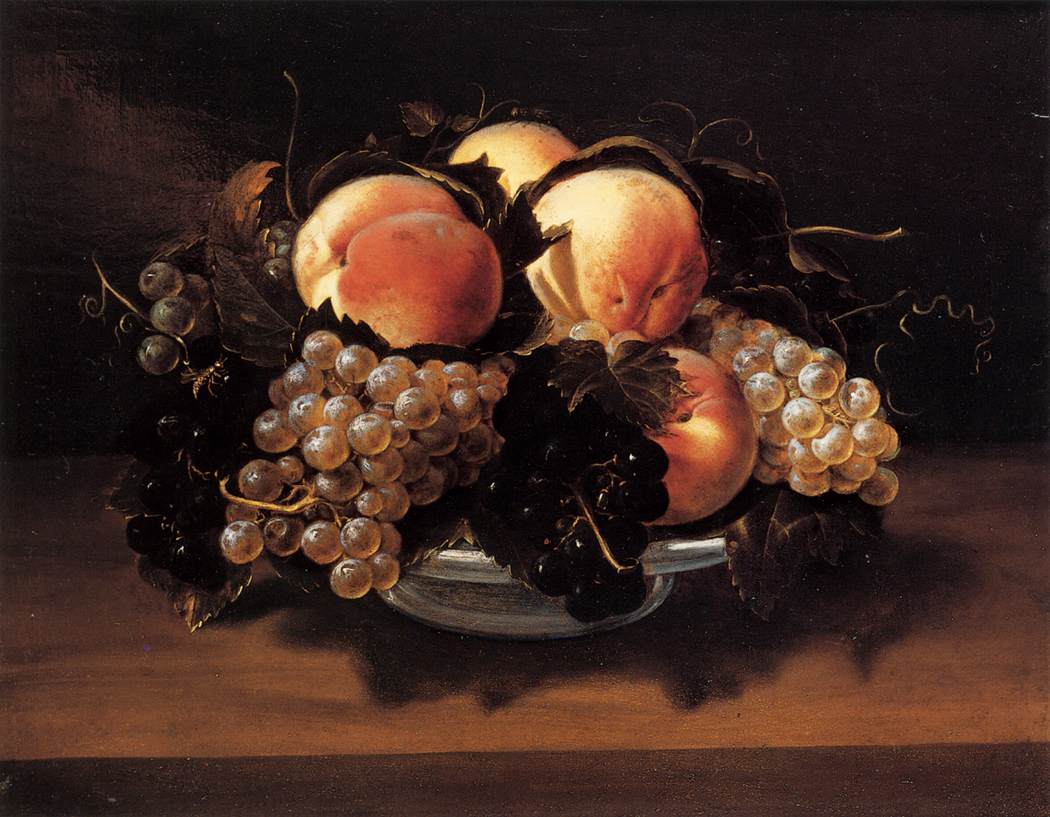} \\
\textcolor{greengt}{\textbf{0.778}} & 0.772 & 0.767 & 0.754 & 0.754  \\ [20pt]

\multicolumn{5}{L{12cm}}{\scriptsize{\textbf{Title}: A Saddled Race Horse Tied to a Fence \linebreak
\textbf{Comment}: Horace Vernet enjoyed royal patronage, one of his earliest commissions was a group of ten paintings depicting Napoleon's horses. These works reveal his indebtedness to the English tradition of horse painting. The present painting was commissioned in Paris in 1828 by Jean Georges Schickler, a member of a German based banking family, who had a passion for horse racing.}} \\
& \\ [-5pt]
\includegraphics[width = 0.18\textwidth, height=60pt]{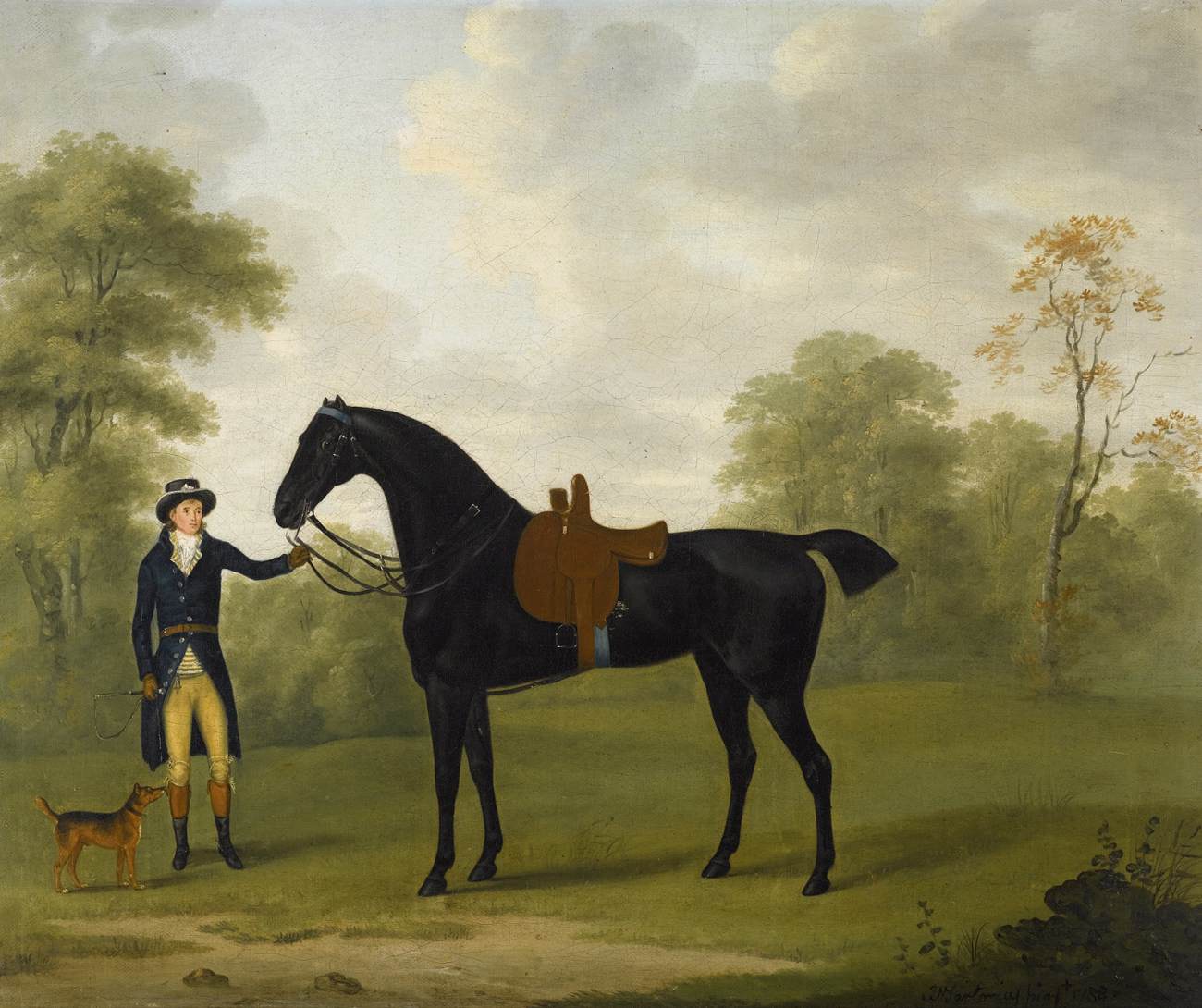} &
\includegraphics[width = 0.18\textwidth, height=60pt, cfbox=greengt 2pt 0pt]{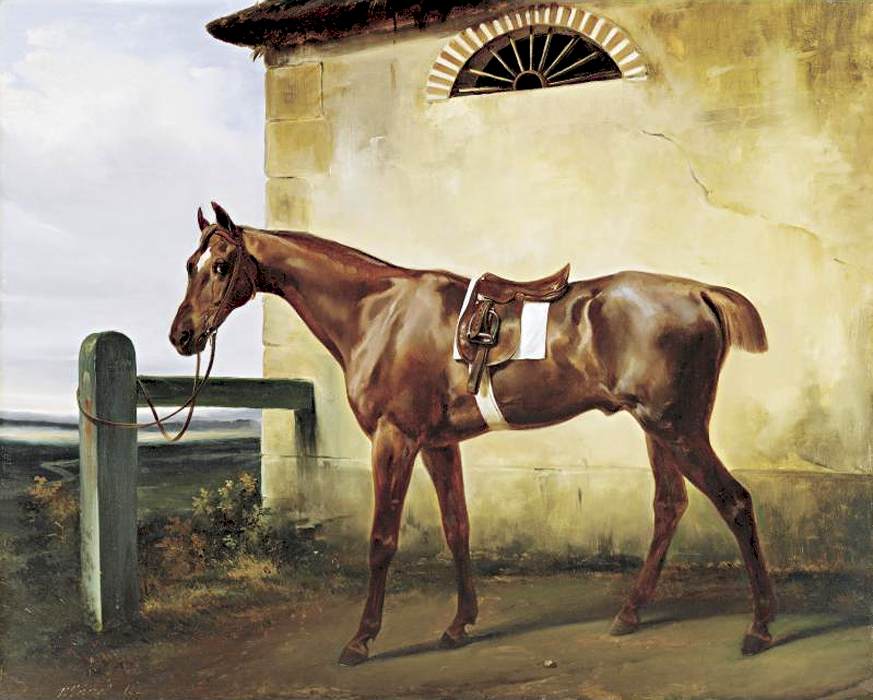} &
\includegraphics[width = 0.18\textwidth, height=60pt]{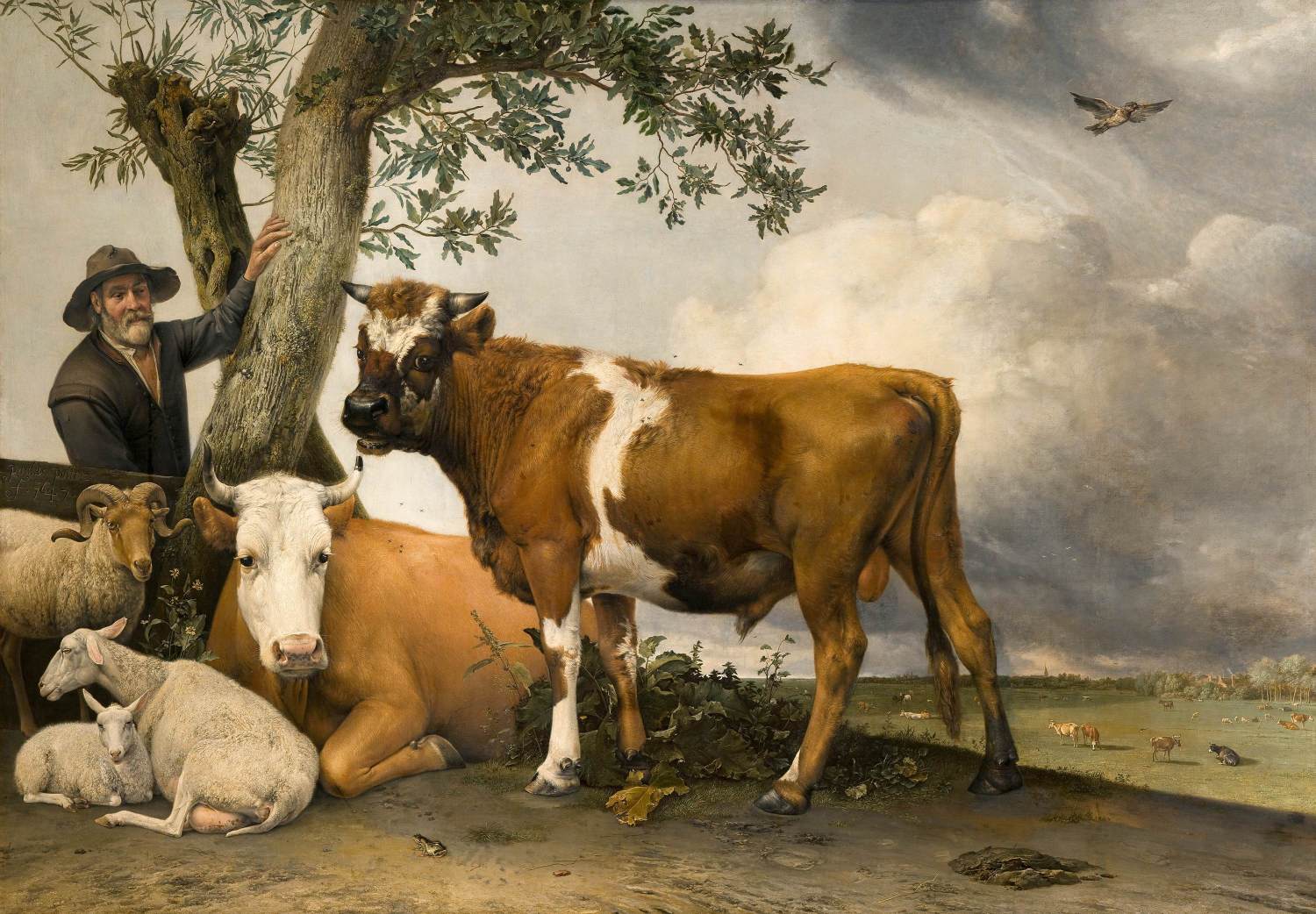} &
\includegraphics[width = 0.18\textwidth, height=60pt]{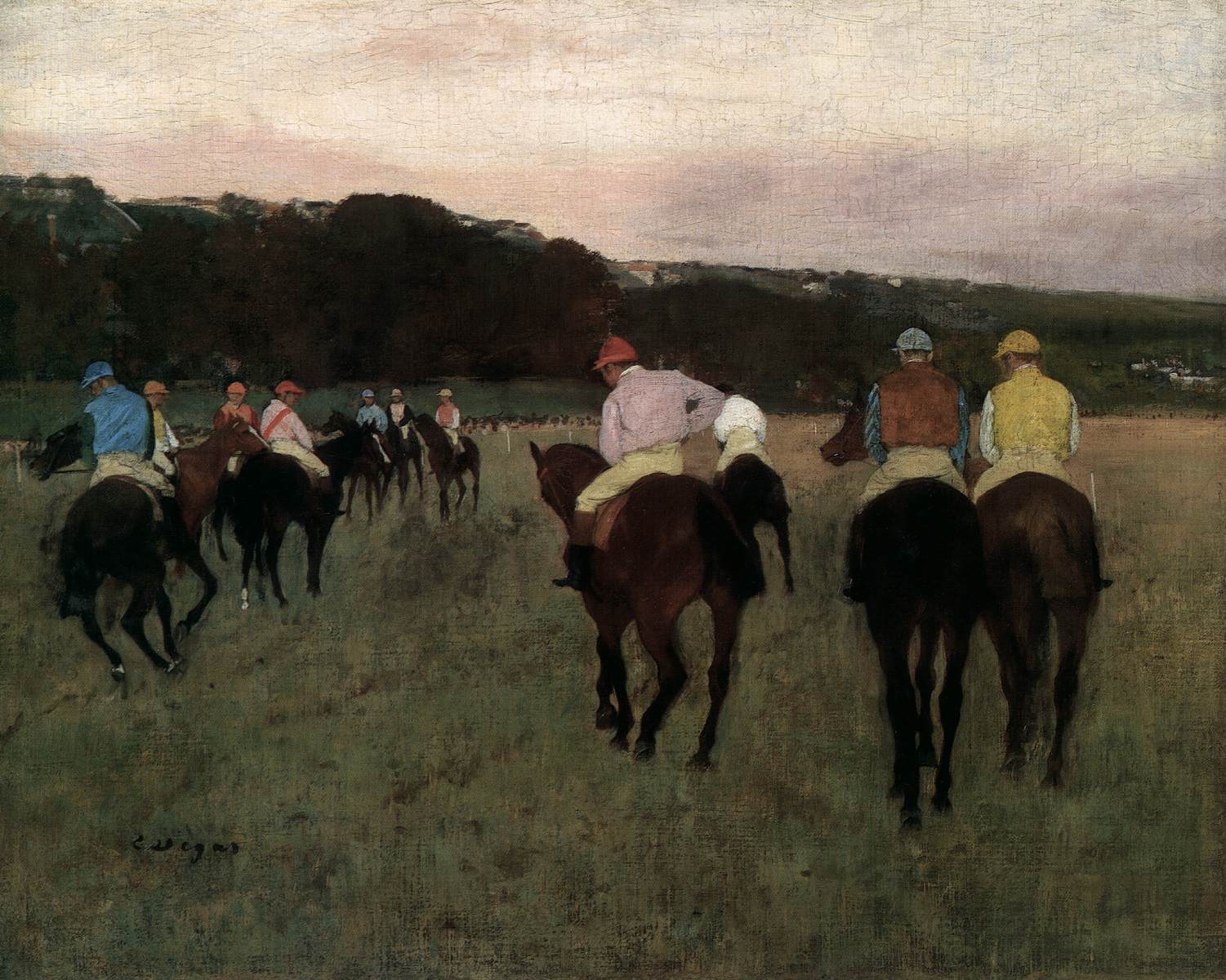} &
\includegraphics[width = 0.18\textwidth, height=60pt]{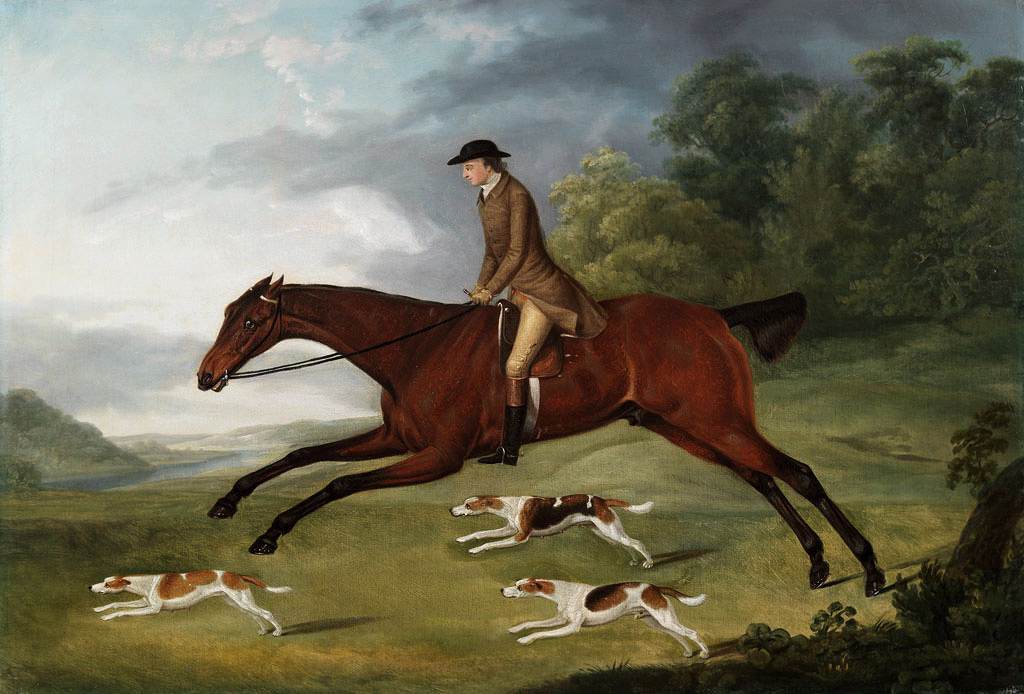} \\
0.755 & \textcolor{greengt}{\textbf{0.732}} & 0.718 & 0.662 & 0.660  \\ [20pt]

\multicolumn{5}{L{12cm}}{\scriptsize{\textbf{Title}: Portrait of a Girl \linebreak
\textbf{Comment}: This painting shows a girl in a yellow dress holding a bouquet of flowers. It is a typical portrait of the artist showing the influence of his teacher, Agnolo Bronzino.}} \\
& \\ [-5pt]
\includegraphics[width = 0.18\textwidth, height=60pt, cfbox=greengt 2pt 0pt]{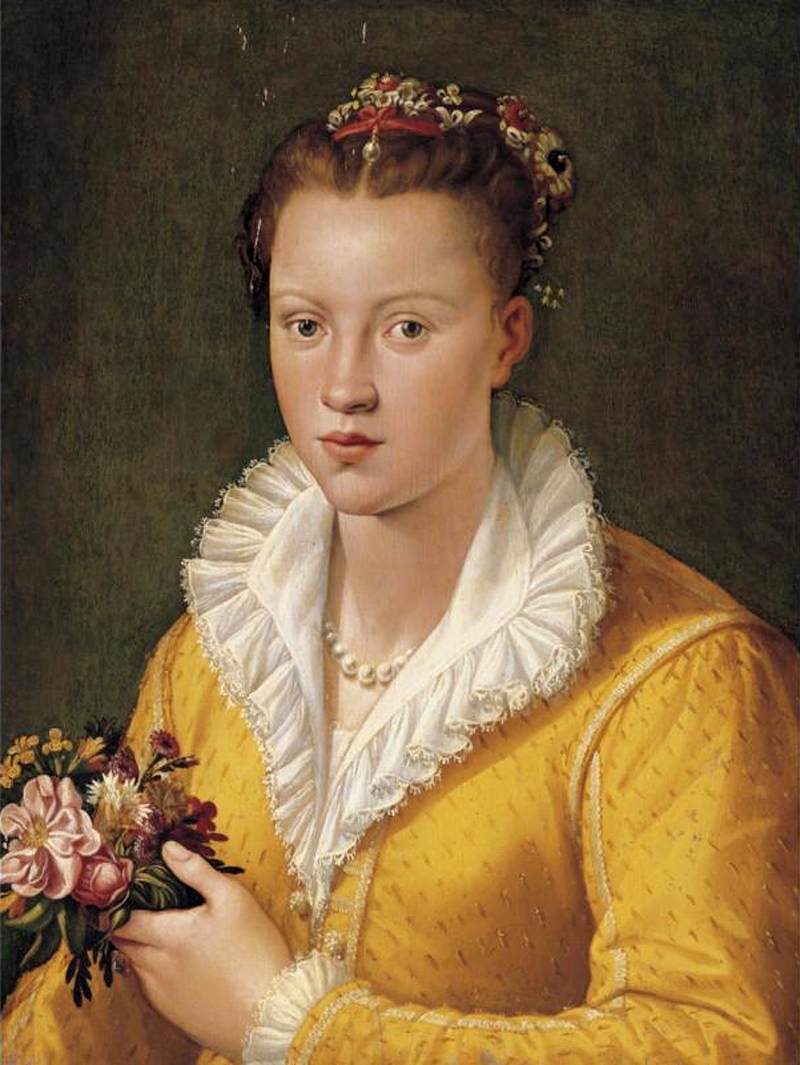} &
\includegraphics[width = 0.18\textwidth, height=60pt]{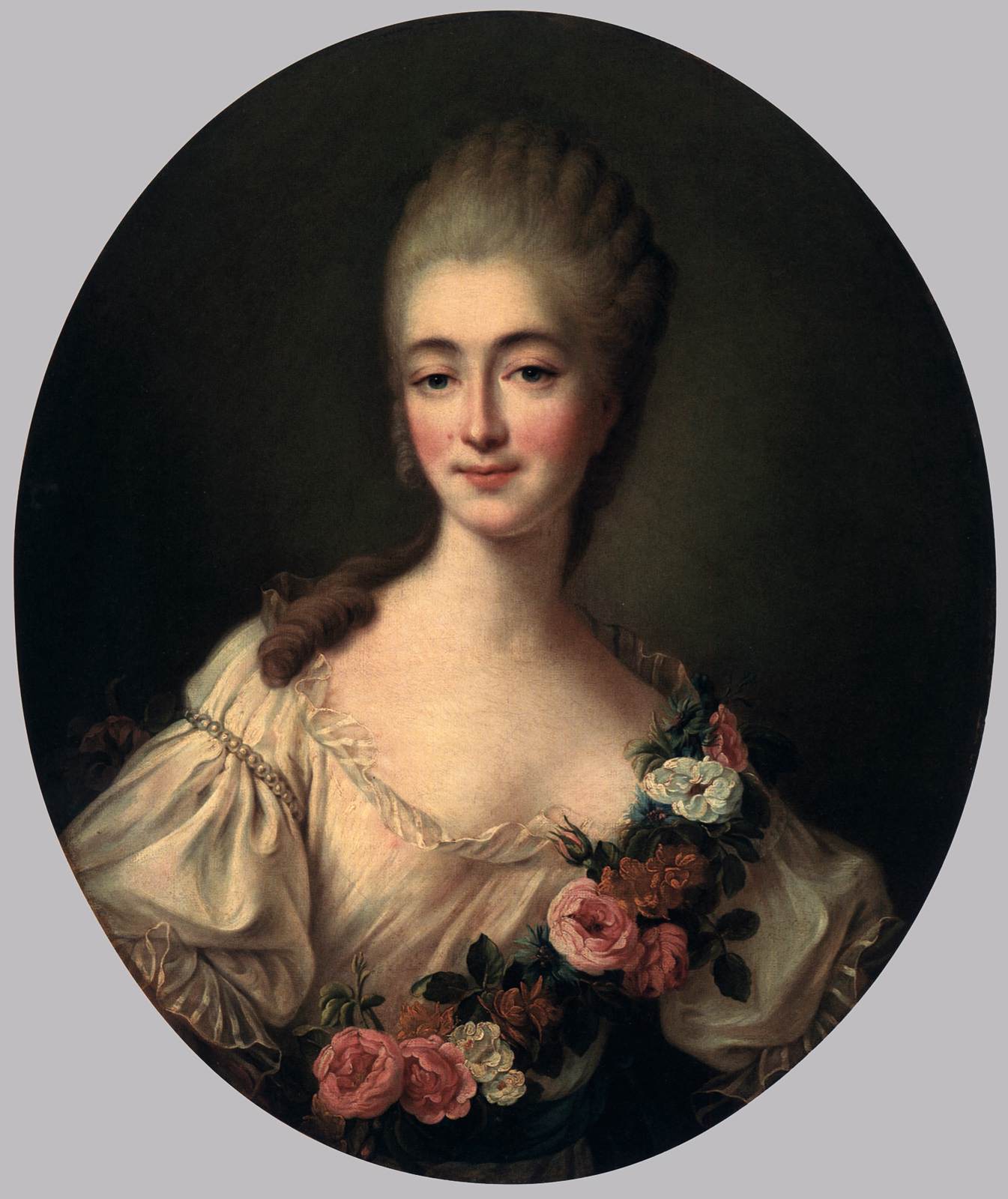} &
\includegraphics[width = 0.18\textwidth, height=60pt]{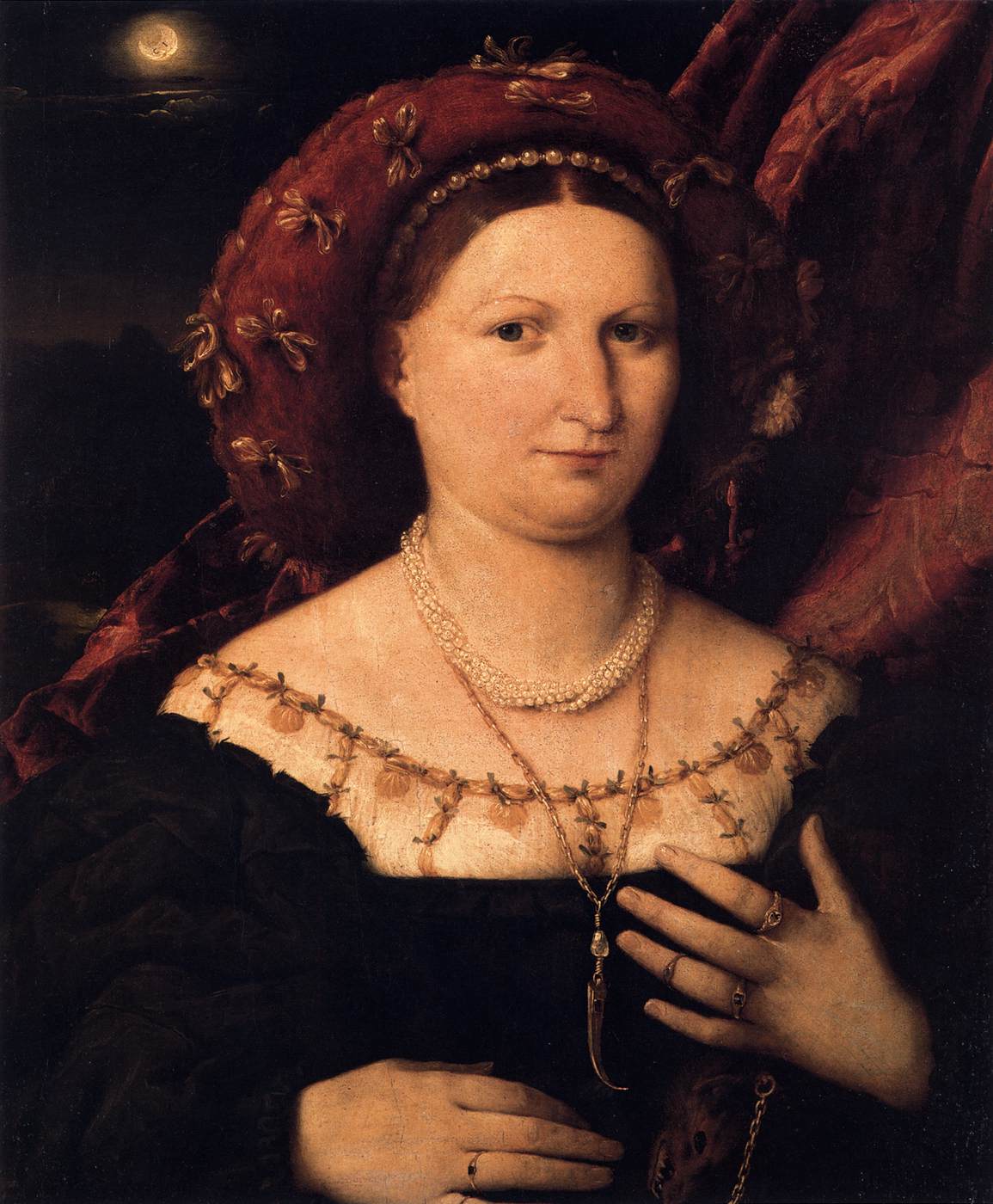} &
\includegraphics[width = 0.18\textwidth, height=60pt]{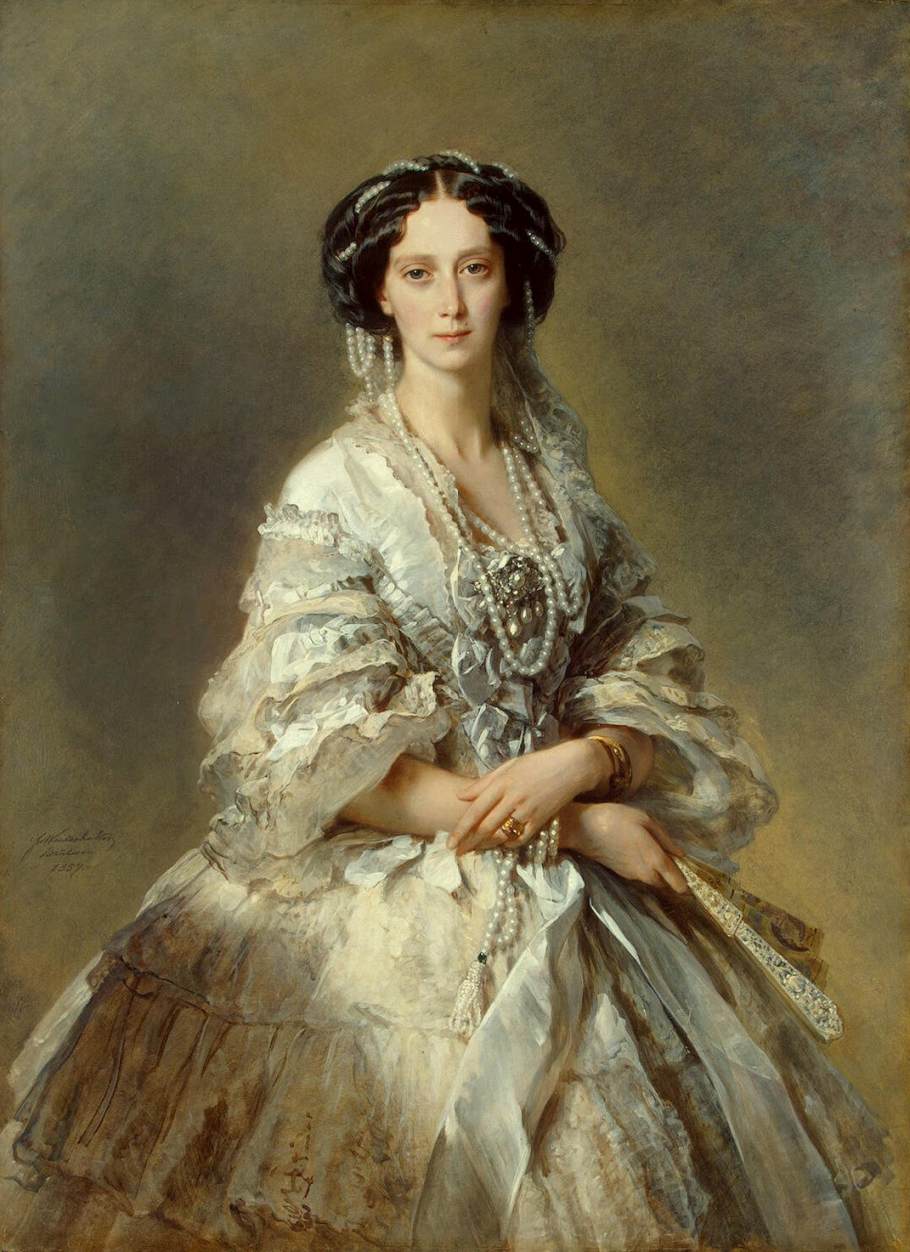} &
\includegraphics[width = 0.18\textwidth, height=60pt]{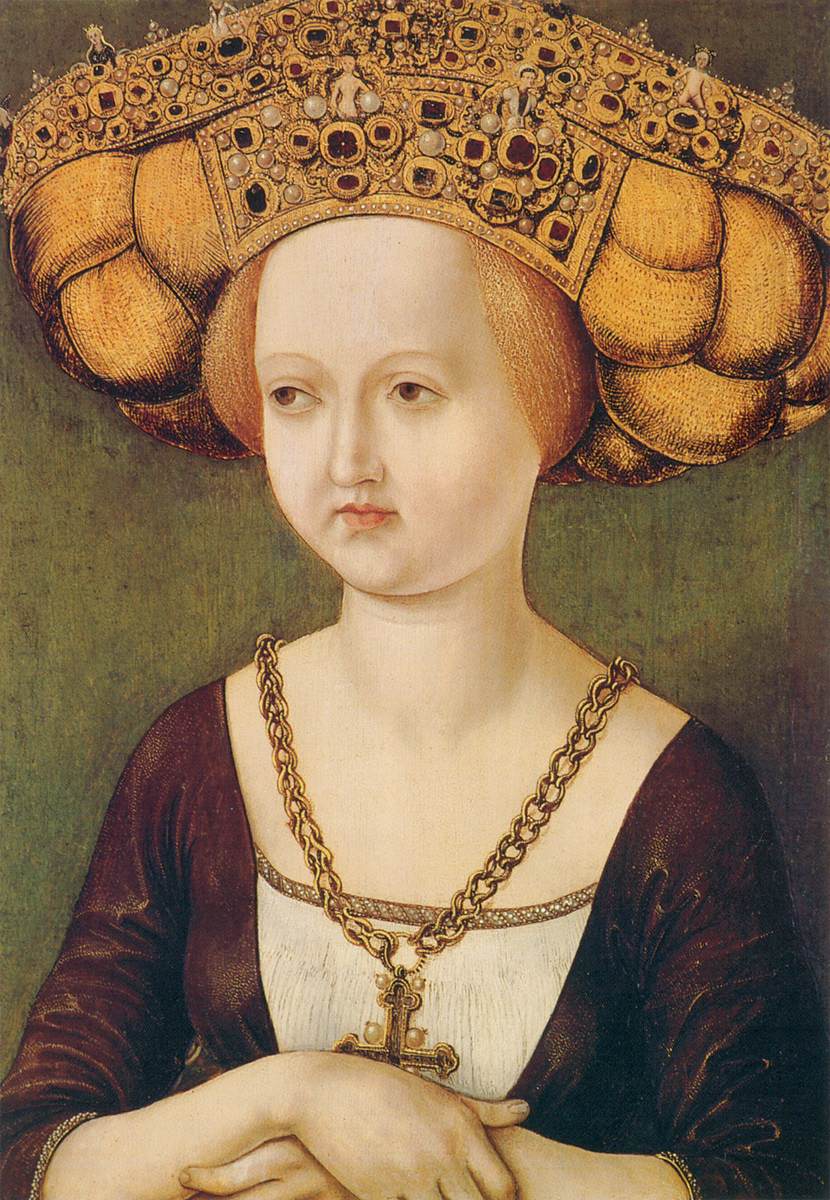} \\ 
\textcolor{greengt}{\textbf{0.870}} & 0.848 & 0.847 & 0.827 & 0.825  \\ [20pt]

\multicolumn{5}{L{12cm}}{\scriptsize{\textbf{Title}: The Kreuzkirche in Dresden \linebreak
\textbf{Comment}: A few years later, during his second stay in Saxony, Bellotto depicted the demolition of this Gothic church. There exists an almost identical version in the Gemldegalerie, Dresden.}} \\
& \\ [-5pt]
\includegraphics[width = 0.18\textwidth, height=60pt, cfbox=greengt 2pt 0pt]{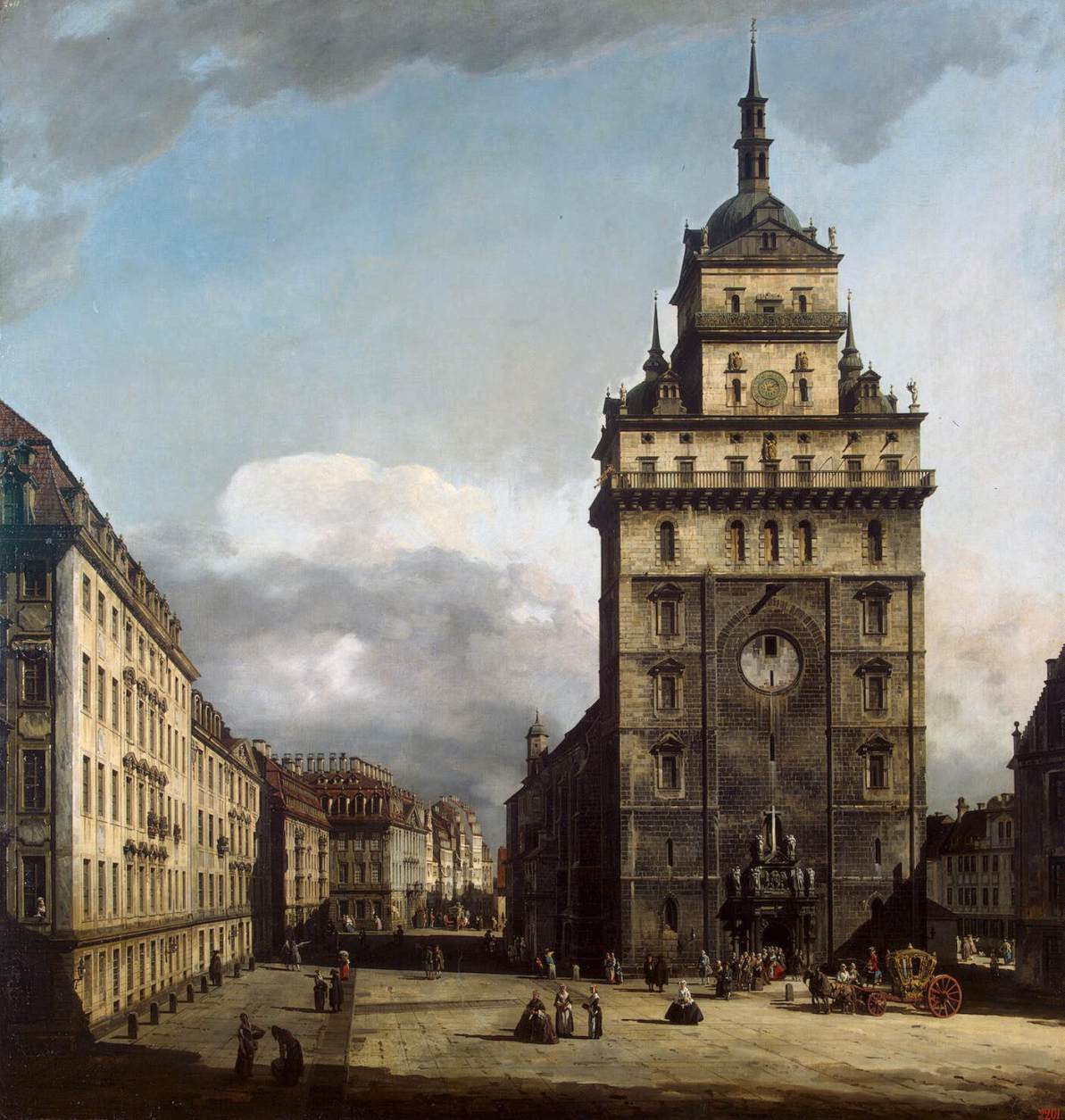} &
\includegraphics[width = 0.18\textwidth, height=60pt]{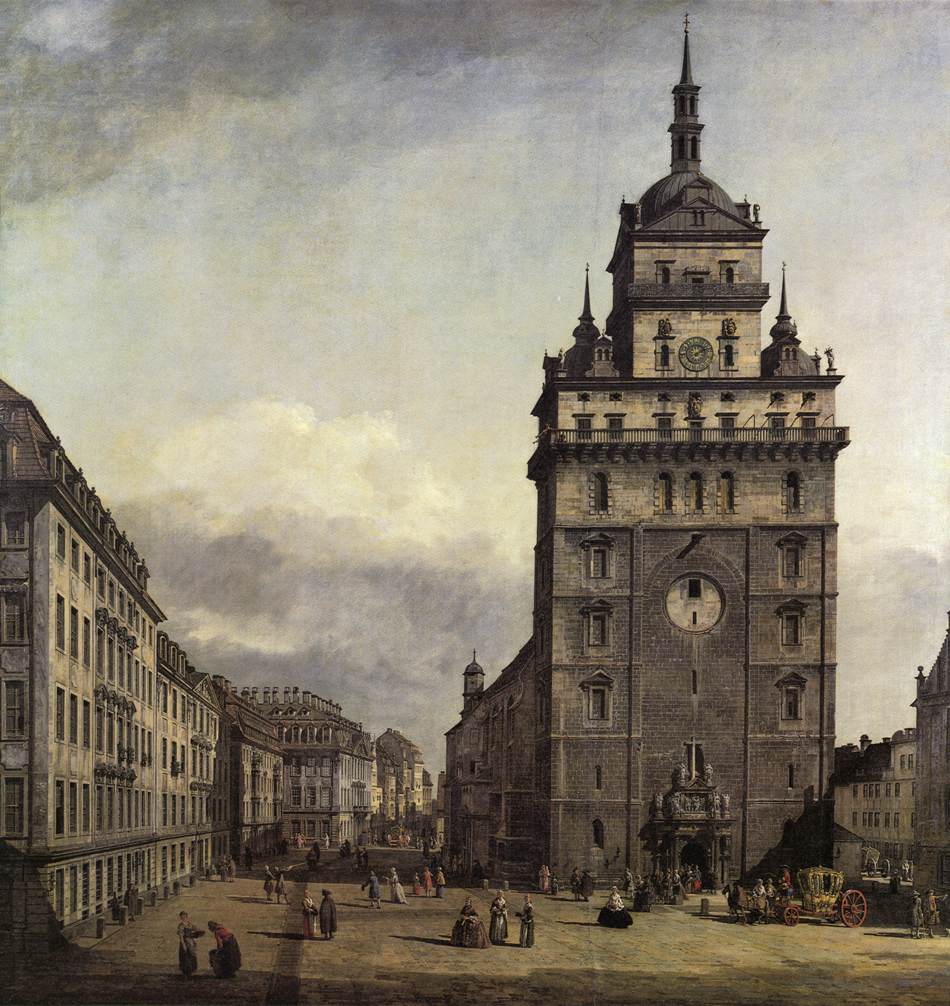} &
\includegraphics[width = 0.18\textwidth, height=60pt]{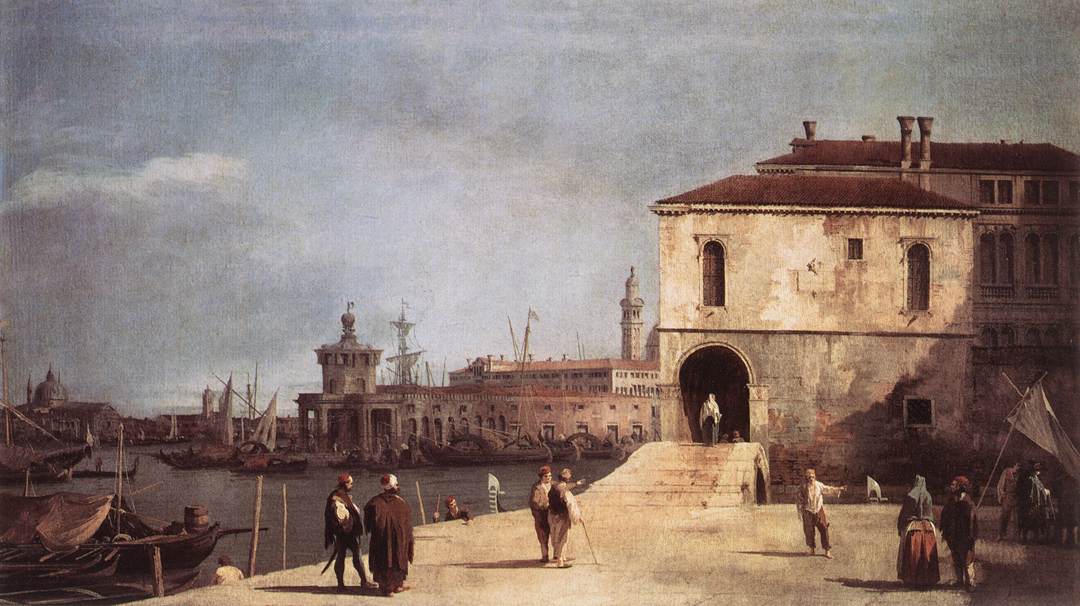} &
\includegraphics[width = 0.18\textwidth, height=60pt]{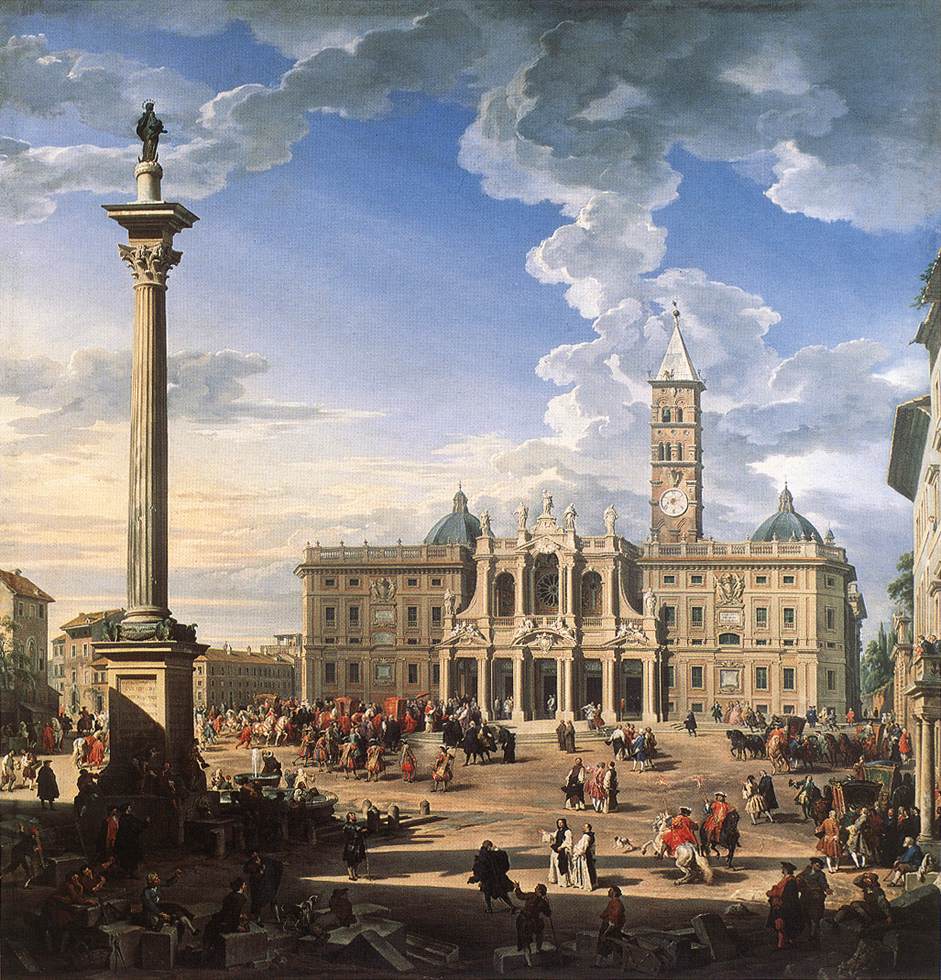} &
\includegraphics[width = 0.18\textwidth, height=60pt]{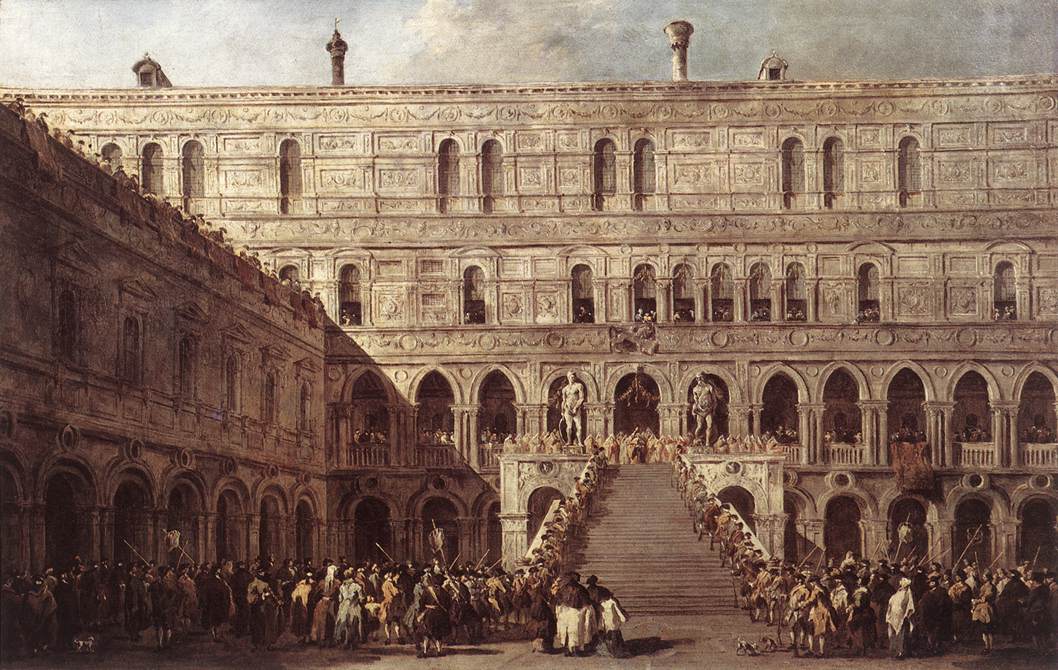} \\
\textcolor{greengt}{\textbf{0.841}} & 0.834 & 0.803 & 0.800 & 0.799  \\ [5pt]

\end{tabular}}
\caption{\textbf{Qualitative positive results}. For each text (i.e. title and comment), the top five ranked images, along with their score, are shown. The ground truth image is highlighted in green.}
\label{fig:qualitativepos}
\end{figure}

\begin{figure}[h!]
\centering
\setlength{\tabcolsep}{3pt}
\begin{tabular}[t]{c c c c c c}

\raisebox{-.15\height}{\includegraphics[width = 0.18\textwidth, height=60pt, cfbox=red 2pt 0pt]{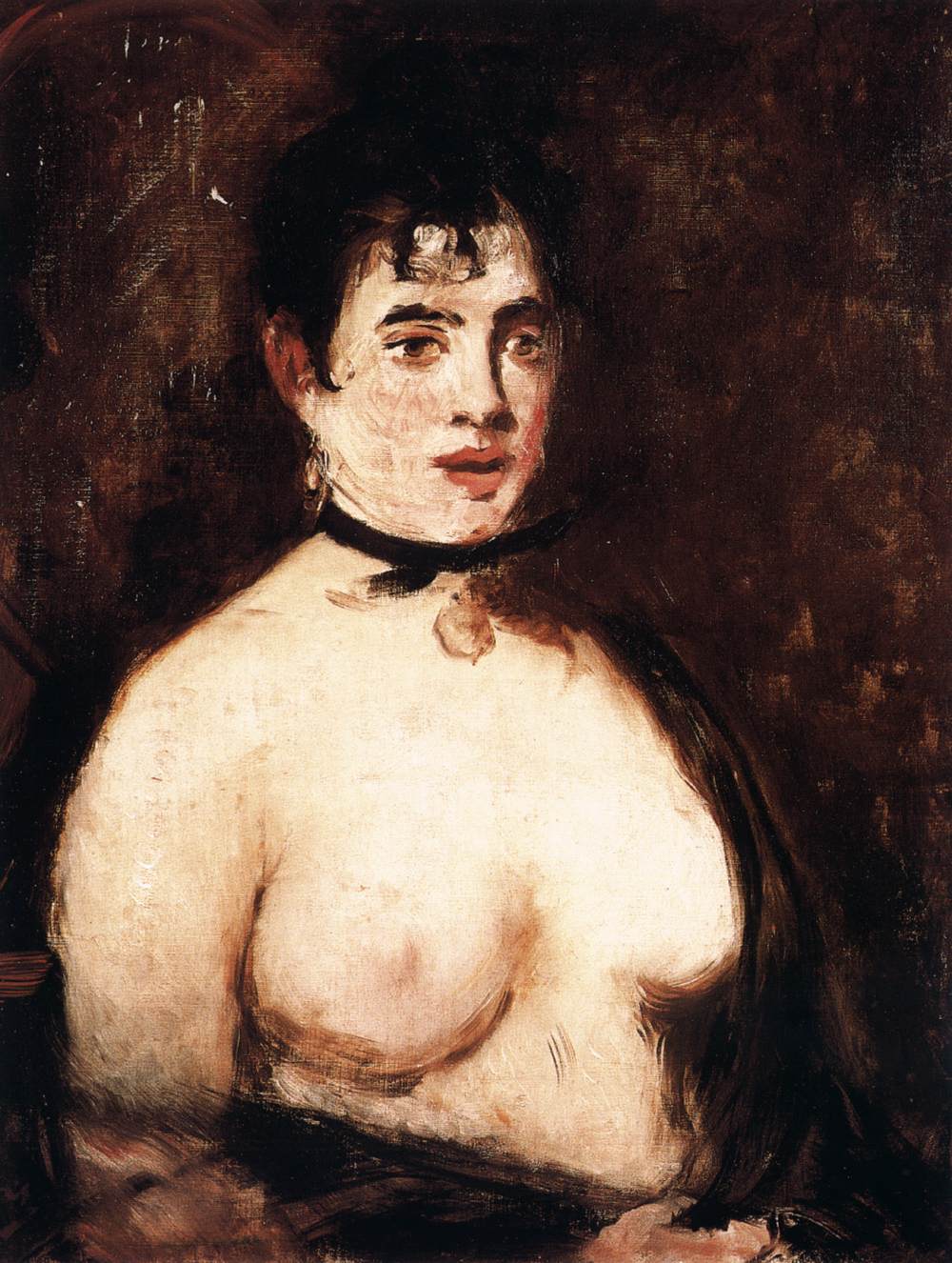}} & \multicolumn{4}{L{10cm}}{\scriptsize{\textbf{Title}: Brunette with Bare Breasts \linebreak
\textbf{Comment}: The 1870s were rich in female models for Manet: the Brunette with Bare Breasts, the Blonde with Bare Breasts and the Sultana testify to it.}} \\
\scriptsize \textcolor{red}{\textbf{ranked 28, 0.445}} & \\
& \\ [-5pt]
\includegraphics[width = 0.19\textwidth, height=60pt]{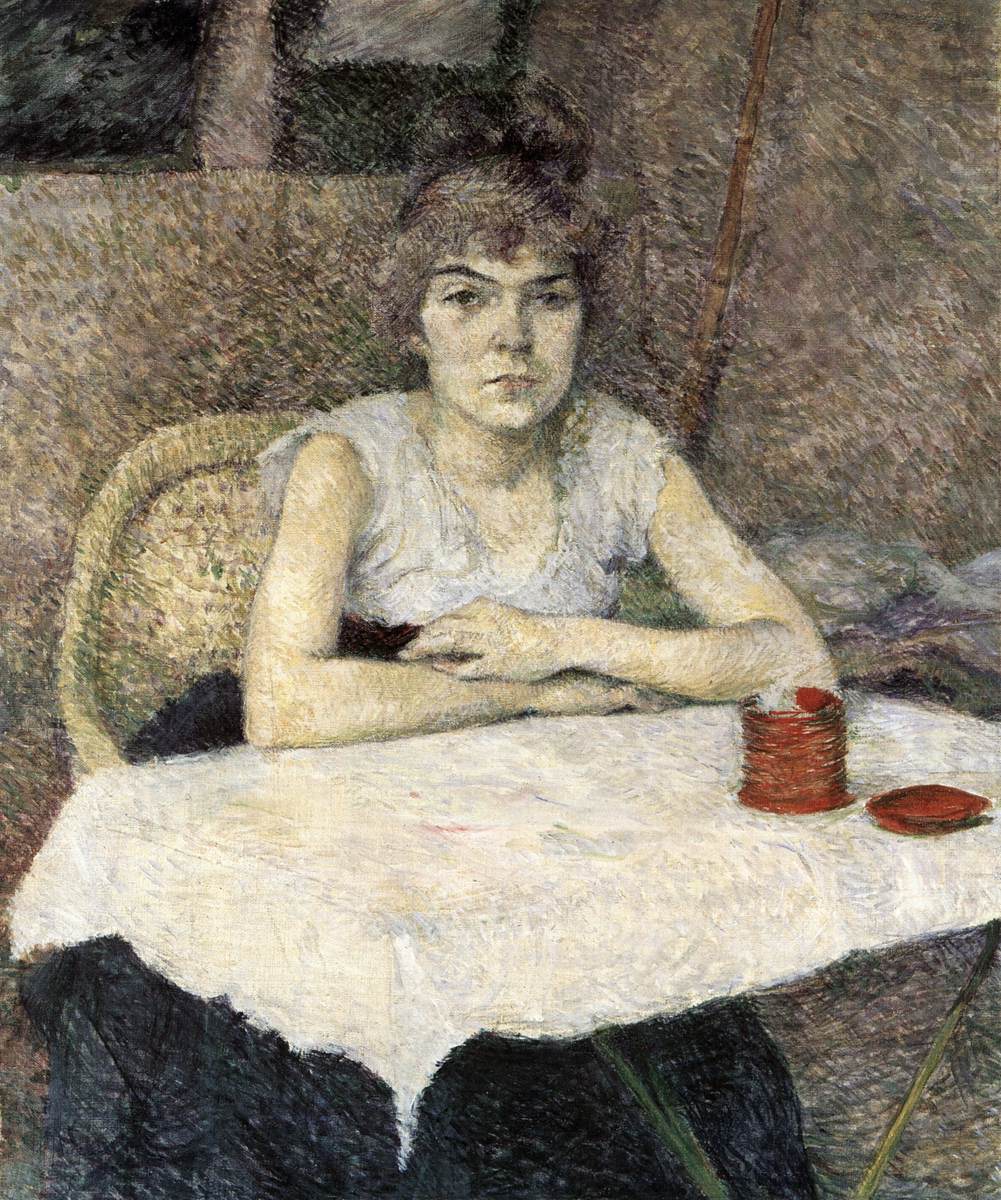} &
\includegraphics[width = 0.19\textwidth, height=60pt]{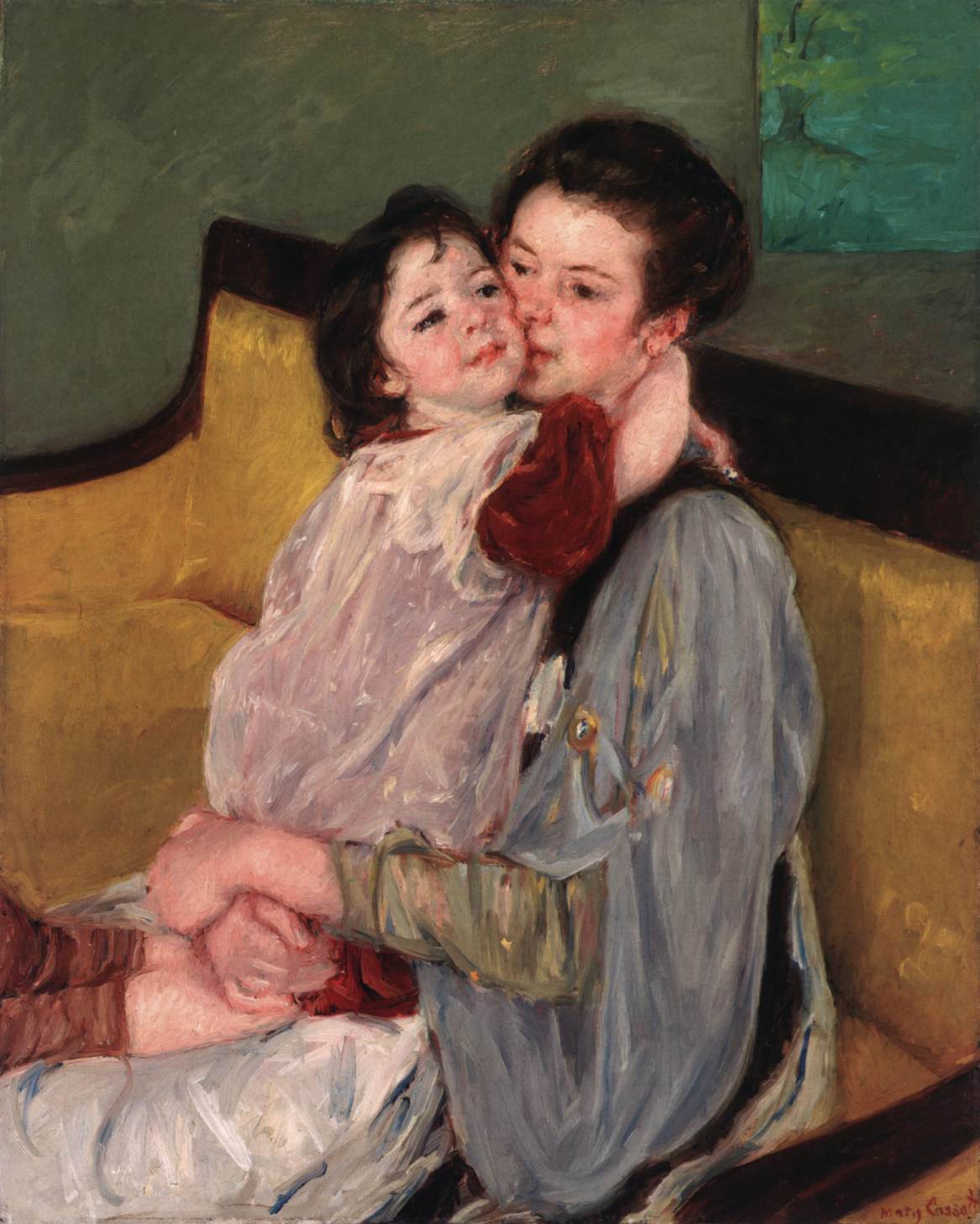} &
\includegraphics[width = 0.19\textwidth, height=60pt]{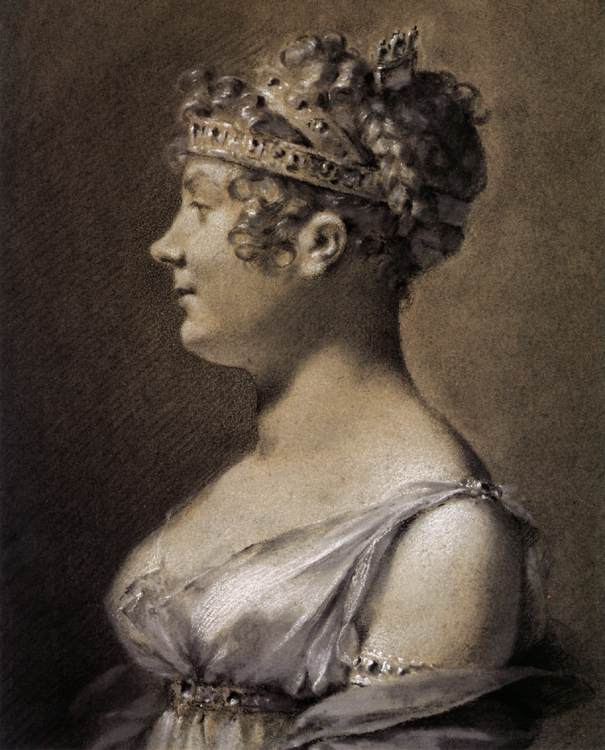} &
\includegraphics[width = 0.19\textwidth, height=60pt]{Images/qualitative/44216-portmari.jpg} &
\includegraphics[width = 0.19\textwidth, height=60pt]{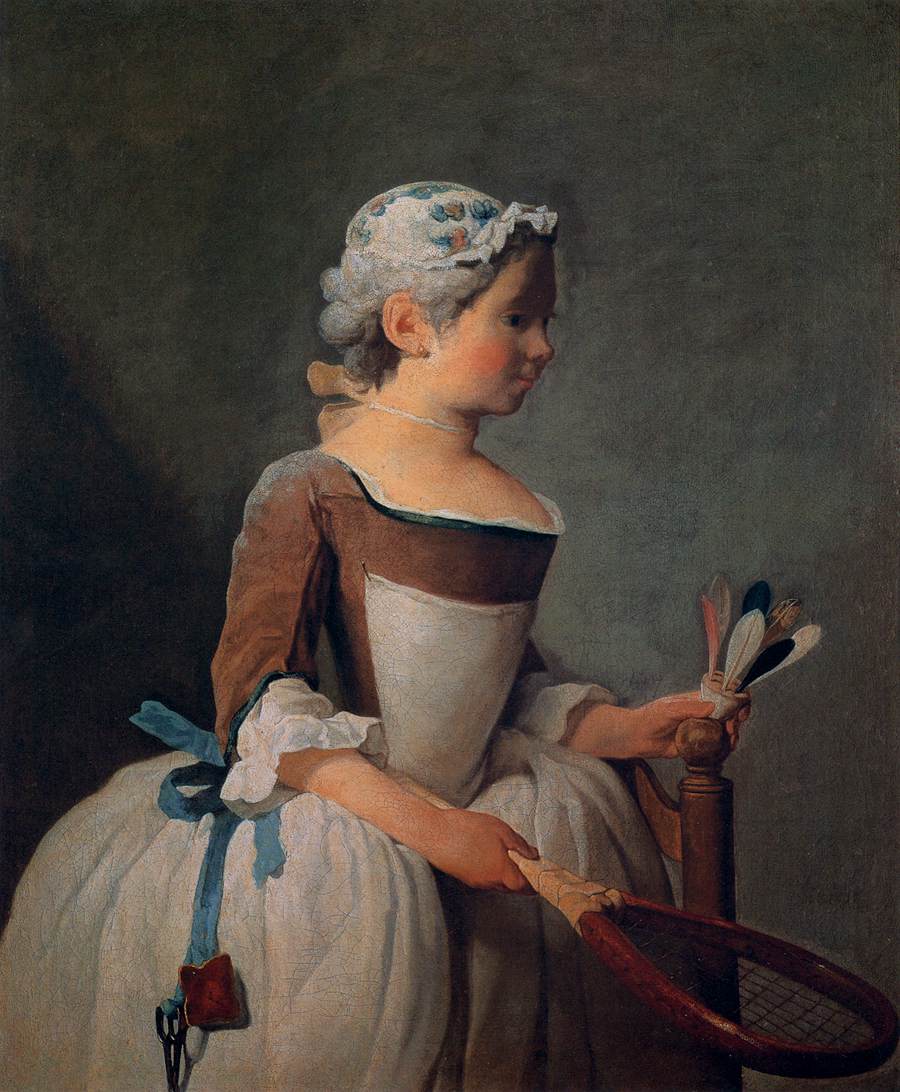} \\[-5pt]
\scriptsize{0.640} & \scriptsize{0.622} & \scriptsize{0.605} & \scriptsize{0.572} & \scriptsize{0.569} \\ [10pt]

\raisebox{-.20\height}{\includegraphics[width = 0.18\textwidth, height=60pt, , cfbox=red 2pt 0pt]{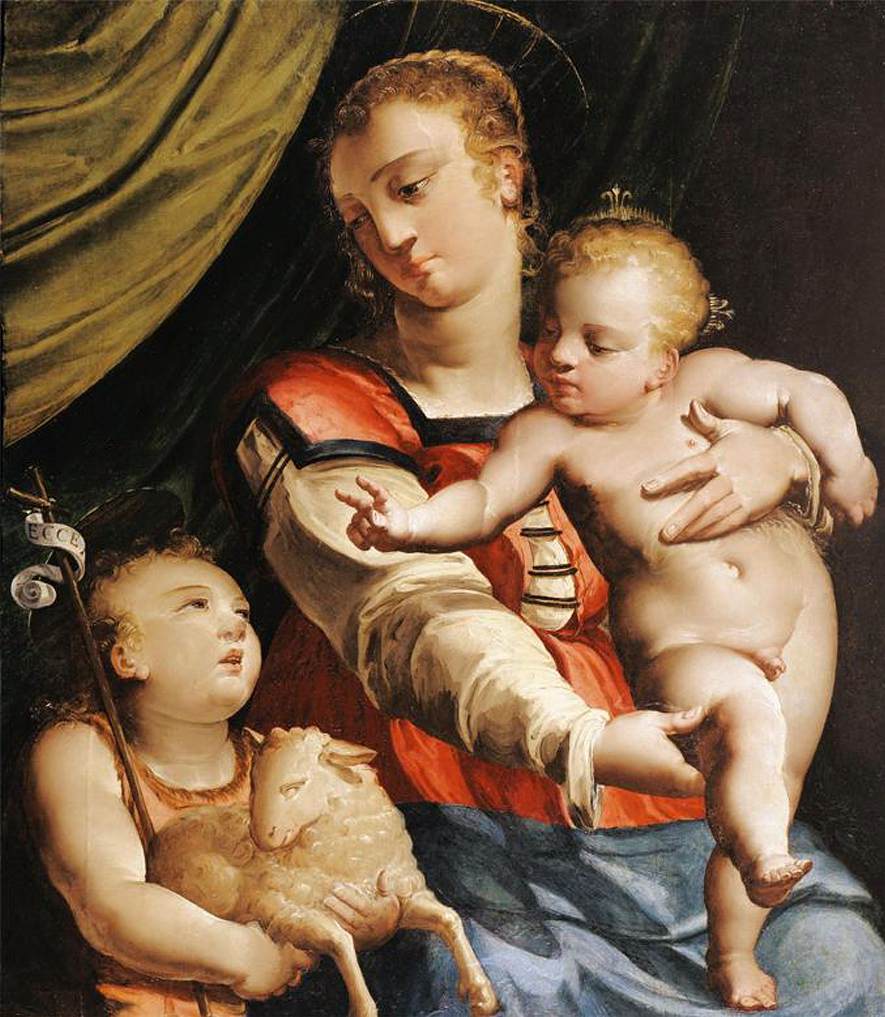}} & \multicolumn{4}{L{10cm}}{\scriptsize{\textbf{Title}: Virgin and Child with the Young St John the Baptist \linebreak
\textbf{Comment}: The stylistic characteristics of this painting, such as rounded faces and narrow, elongated eyes seem to be a general reflection of the foreign presence in Genoese painting at this time.}} \\
\scriptsize \textcolor{red}{\textbf{ranked 17, 0.690}} & \\
& \\ [-5pt]
\includegraphics[width = 0.19\textwidth, height=60pt]{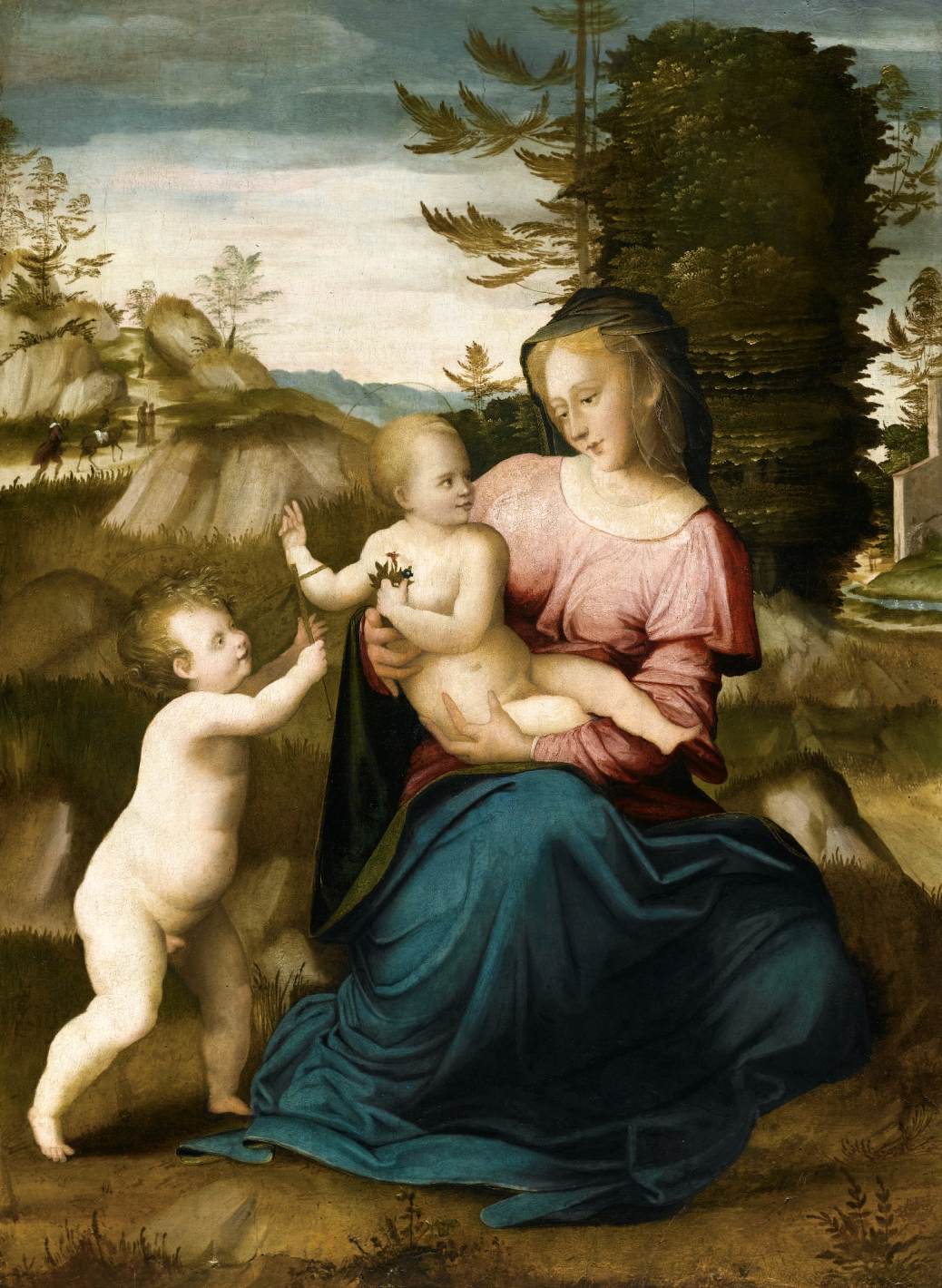} &
\includegraphics[width = 0.19\textwidth, height=60pt]{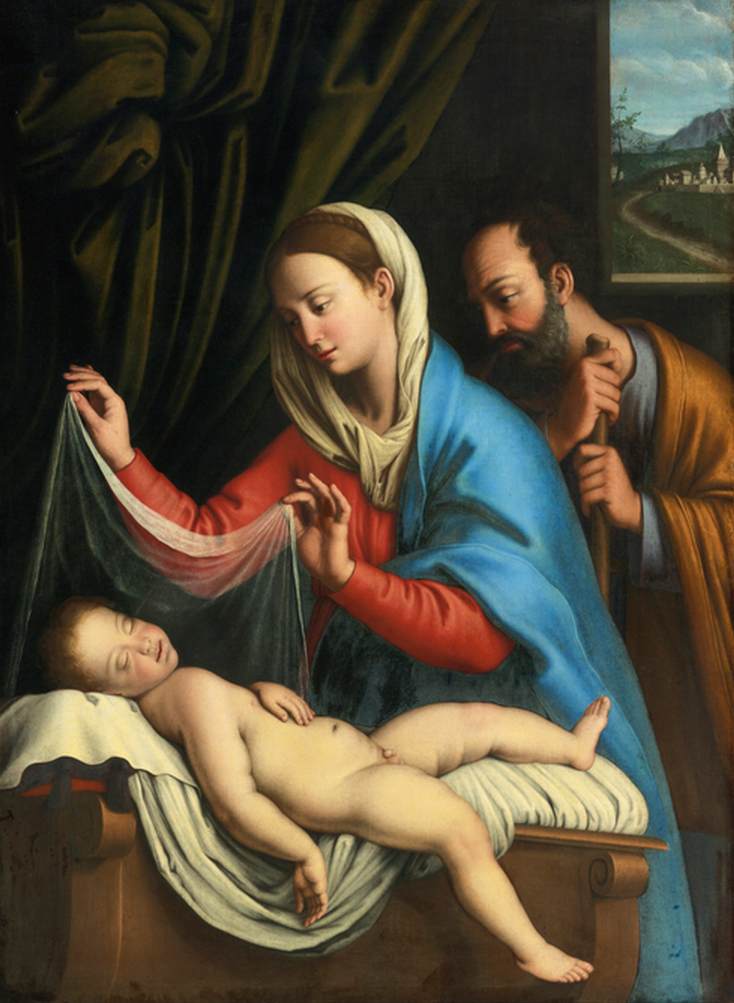} &
\includegraphics[width = 0.19\textwidth, height=60pt]{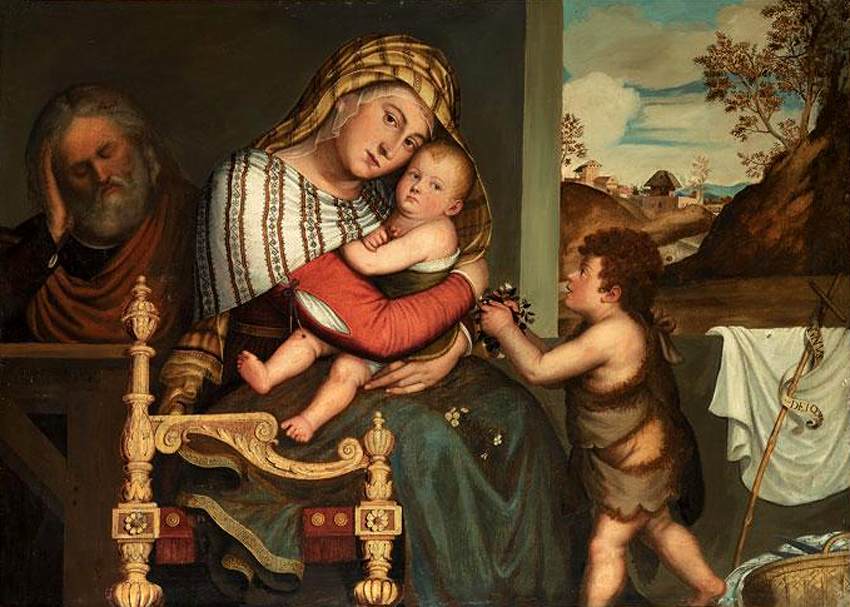} &
\includegraphics[width = 0.19\textwidth, height=60pt]{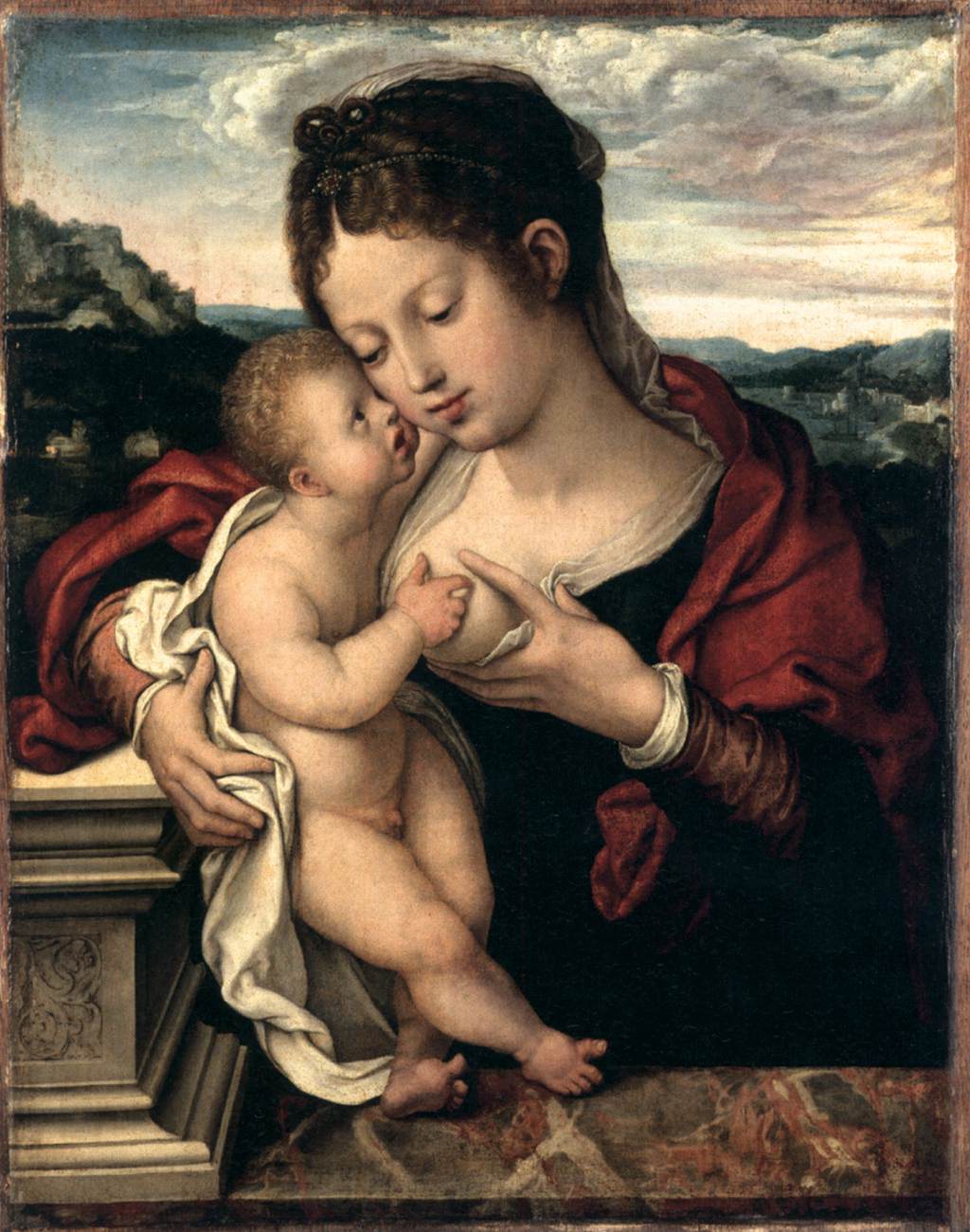} &
\includegraphics[width = 0.19\textwidth, height=60pt]{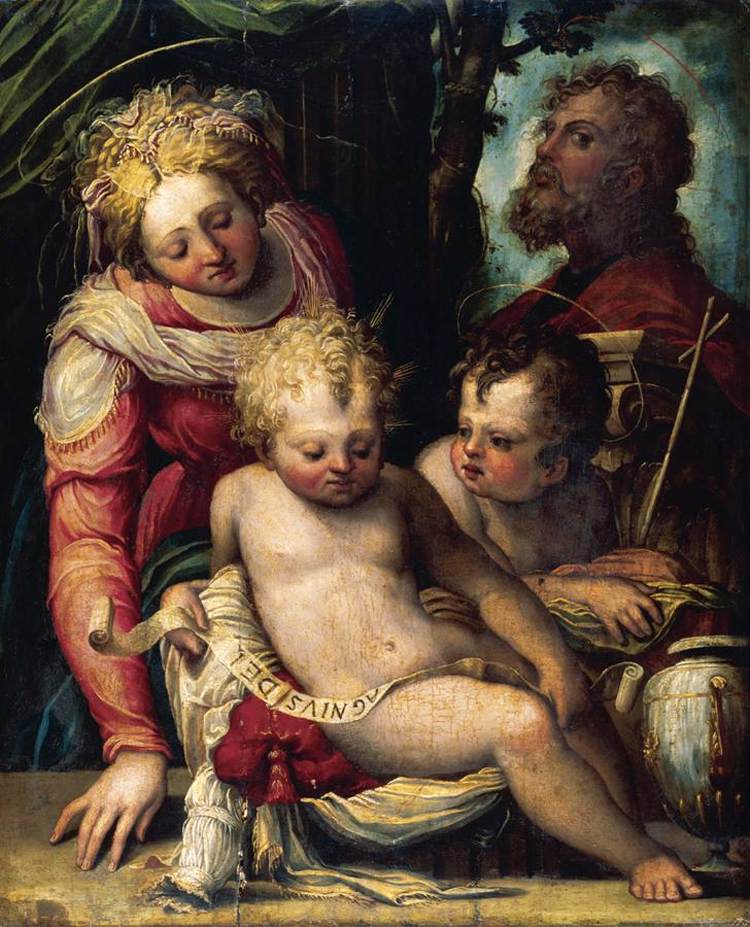} \\[-5pt]
\scriptsize{0.754} & \scriptsize{0.751} & \scriptsize{0.730} & \scriptsize{0.727} & \scriptsize{0.721} \\
\end{tabular}
\caption{\textbf{Qualitative negative result}. For each text, the ground truth image is shown next to it, along with its ranking position and score. Below, the five top ranked images.}
\label{fig:qualitativeneg}
\end{figure}

Finally, we compare the three proposed multi-modal transformation models in the Text2Art challenge: CCA, CML and AMD. For the AMD approach, we use four different attributes to inform the model: Type (AMD\scriptsize{T}\normalsize), TimeFrame (AMD\scriptsize{TF}\normalsize), School (AMD\scriptsize{S}\normalsize) and Author (AMD\scriptsize{A}\normalsize). ResNet50 is used to encode visual features. Results are shown in Table \ref{tab:models}. Random ranking results are provided as reference. Overall, the best performance is achieved with the CML model and bag-of-words encodings. CCA achieves the worst results among all the models, which suggests that linear transformations are not able to adjust properly to the task. Surprisingly, adding extra information in the AMD models does not lead to further improvement over the CML approach. We suspect that this might be due to the unbalanced number of samples within the classes of the dataset. Qualitative results of the CML model with ResNet50 and bag-of-words encodings are shown in Figures \ref{fig:qualitativepos} and \ref{fig:qualitativeneg}. In the positive examples (Figure \ref{fig:qualitativepos}), not only the ground truth painting is ranked within the top five returned images, but also all the images within the top five are semantically similar to the query text. In the unsuccessful examples (Figure \ref{fig:qualitativeneg}), although the ground truth image is not ranked in the top positions of the list, the algorithm returns images that are semantically  meaningful to fragments of the text, which indicates how challenging the task is.

\begin{table}
\setlength{\tabcolsep}{5pt}
\caption{\textbf{Multi-modal transformation models.} Comparison between different multi-modal transformation models in the Text2Art challenge.}
\centering
\resizebox{\textwidth}{!}{%
\begin{tabular}{ c c c c c c c c c c c c c c}

\Xhline{2\arrayrulewidth}

\multicolumn{3}{c}{\textbf{Technique}} & & \multicolumn{4}{c}{\textbf{Text-to-Image}} & & \multicolumn{4}{c}{\textbf{Image-to-Text}} \\ \cline{1-3} \cline{5-8} \cline{10-13}

\textbf{Model} & \textbf{Com} & \textbf{Att} & & \textbf{R@1} & \textbf{R@5} & \textbf{R@10} & \textbf{MR} & & \textbf{R@1} & \textbf{R@5} & \textbf{R@10} & \textbf{MR} \\ \hline 

Random & - & - & & 0.0008 & 0.004 & 0.009 & 539 & & 0.0008 & 0.004 & 0.009 & 539 \\ 

CCA & MLP\scriptsize{c} & MLP\scriptsize{a} & & 0.117 & 0.283 & 0.377 & 25 & & 0.131 & 0.279 & 0.355 & 26\\

CML & BOW\scriptsize{c} & BOW\scriptsize{a} & & \textbf{0.144} & \textbf{0.332} & \textbf{0.454} & \textbf{14} & & 0.138 & \textbf{0.327} & \textbf{0.457} & \textbf{14} \\
CML & MLP\scriptsize{c} & MLP\scriptsize{a} & & 0.137 & 0.306 & 0.432 & 16 & & \textbf{0.140} & 0.317 & 0.436 & 15 \\

AMD\scriptsize{T} & MLP\scriptsize{c} & MLP\scriptsize{a} & & 0.114 & 0.304 & 0.398 & 17 & & 0.125 & 0.280 & 0.398 & 16  \\
AMD\scriptsize{TF} & MLP\scriptsize{c} & MLP\scriptsize{a} & & 0.117 & 0.297 & 0.389 & 20 & & 0.123 & 0.298 & 0.413 & 17 \\
AMD\scriptsize{S} & MLP\scriptsize{c} & MLP\scriptsize{a} & & 0.103 & 0.283 & 0.401 & 19 & & 0.118 & 0.298 & 0.423 & 16 \\
AMD\scriptsize{A} & MLP\scriptsize{c} & MLP\scriptsize{a} & & 0.131 & 0.303 & 0.418 & 17 & & 0.120 & 0.302 & 0.428 & 16\\ \hline

\Xhline{2\arrayrulewidth}
\end{tabular}}
\label{tab:models}
\end{table}

\subsection{Human Evaluation}
We design a task in Amazon Mechanical Turk\footnote{https://www.mturk.com/} for testing  human performance in the Text2Art challenge. For a given artistic text, which includes comment, title, author, type, school and timeframe, human evaluators are asked to choose the most appropriate painting from a pool of ten images. The task has two different levels: \textit{easy}, in which the pool of images is chosen randomly from all the paintings in test set, and \textit{difficult}, in which the ten images in the pool share the same attribute type (i.e. portraits, landscapes, etc.). For each level, evaluators are asked to perform the task in 100 artistic texts. Accuracy is measured as the ratio of correct answers over the total number of answers. Results are shown in Table \ref{tab:humaneasy}. Although human accuracy is considerable high, reaching 88.9\% in the easiest set, there is a drop in performance in the difficult level, mostly because images from the same type contain more similar comments than images from different types. We evaluate a CCA and a CML model in the same data split as humans. The CML model with bag-of-words and ResNet50 is able to find the relevant image in the 75\% of the samples in the easy set and in the 62\% of the cases in the difficult task. There is around ten points of difference between CML model and the human evaluation, which suggests that, although there is still room for improvement, meaningful art representations are being obtained.

\begin{table}[t]
\setlength{\tabcolsep}{4pt}
\caption{\textbf{Human Evaluation}. Evaluation in both the easy and the difficult sets.}
\centering
\resizebox{\textwidth}{!}{%
\begin{tabular}{ c c c c c c c c c c c  c}
\Xhline{2\arrayrulewidth}
& \multicolumn{4}{c}{\textbf{Technique}} & & \multicolumn{6}{c}{\textbf{Text-to-Image}} \\ \cline{2-5} \cline{7-12}
& \textbf{Model} & \textbf{Img} & \textbf{Com} & \textbf{Att} & & \textbf{Land} & \textbf{Relig} & \textbf{Myth} & \textbf{Genre} & \textbf{Port} & \textbf{Total} \\ 
\hline
 \parbox[t]{2pt}{\multirow{3}{*}{\rotatebox[origin=c]{90}{Easy}}}
& CCA & ResNet152 & MLP\scriptsize{c} & MLP\scriptsize{a} & & 0.708 & 0.609 & 0.571 & 0.714 & 0.615 & 0.650  \\ 
& CML & ResNet50 & BOW\scriptsize{c} & BOW\scriptsize{a} & & 0.917 & 0.683 & 0.714 & 1 & 0.538 & 0.750  \\ 
& Human & - & - & - & & 0.918 & 0.795 & 0.864 & 1 & 1 & 0.889 \\
\hline
 \parbox[t]{2pt}{\multirow{3}{*}{\rotatebox[origin=c]{90}{Diff.}}}
& CCA & ResNet152 & MLP\scriptsize{c} & MLP\scriptsize{a} & & 0.600 & 0.525 & 0.400 & 0.300 & 0.400 & 0.470 \\
& CML & ResNet50 & BOW\scriptsize{c} & BOW\scriptsize{a} & & 0.500 & 0.875 & 0.600 & 0.200 & 0.500 & 0.620  \\ 
& Human & - & - & - &  & 0.579 & 0.744 & 0.714 & 0.720 & 0.674 & 0.714 \\
\Xhline{2\arrayrulewidth}
\end{tabular}}
\label{tab:humaneasy}
\end{table}

\section{Conclusions}
\label{sec:conclusions}
We presented the SemArt dataset, the first collection of fine-art images with attributes and artistic comments for semantic art understanding. In SemArt, comments describe artistic information of the painting, such as content, techniques or context. We designed the Text2Art challenge to evaluate semantic art understanding as a multi-modal retrieval task, whereby given an artistic text (or image), a relevant image (or text) is found. We proposed several models to address the challenge. We showed that for visual encoding, ResNets perform the best. For textual encoding, recurrent models performed worse than multi-layer preceptron or bag-of-words. We projected the visual and textual encodings into a common multi-modal space using several methods, the one with the best results being a neural network trained with cosine margin loss. Experiments with human evaluators showed that current approaches are not able to reach human levels of art understanding yet, although meaningful representations for semantic art understanding are being learnt.

\clearpage

\bibliographystyle{splncs04}
\bibliography{egbib}

\end{document}